\documentclass{article}

\usepackage{PRIMEarxiv}
\usepackage[utf8]{inputenc}
\usepackage[T1]{fontenc}
\usepackage[numbers,sort&compress]{natbib}
\usepackage{hyperref}
\usepackage{url}
\usepackage{booktabs}
\usepackage{amsmath}
\usepackage{amssymb}
\usepackage{amsfonts}
\usepackage{graphicx}
\usepackage[dvipsnames]{xcolor}
\usepackage{subcaption}
\usepackage{float}
\hypersetup{hypertexnames=false}

\graphicspath{{media/}}

\newcommand{\bolek}{\textsc{Bolek}}

\title{\bolek: A Multimodal Language Model \\ for Molecular Reasoning}

\author{
  \bfseries
  Frederic Grabowski$^{1,}$\thanks{Equal contribution.} ,
  Jacek Szczerbiński$^{1,}$\footnotemark[2] ,
  Maciej Jaśkowski$^{1}$,
  Kalina Jasińska-Kobus$^{1}$ \\
  \bfseries
  Paweł Dąbrowski-Tumański$^{1,2}$,
  Maciej Wiśniewski$^{1,3}$,
  Tomasz Jetka$^{1}$,
  Bartosz Topolski$^{1}$ \\
  \normalfont $^{1}$Ingenix.ai, Warsaw, Poland \\
  \normalfont $^{2}$Laboratory of Bioinformatics and Computational Genomics, \\
  \normalfont Warsaw University of Technology, Warsaw, Poland \\
  \normalfont $^{3}$Warsaw University of Technology, Warsaw, Poland \\
  \normalfont \texttt{tomasz.jetka@ingenix.ai}
}

\begin{document}
\maketitle
\setcounter{footnote}{0}

\begin{abstract}
AI models and platforms for molecular science underpin high-stakes applications in drug discovery, yet the systems delivering them are largely opaque: they expose either a score and a binary answer with no rationale, or fluent prose rarely anchored in molecular structure.
We introduce \bolek{}, a compact multimodal language model that injects a molecular embedding into an instruction-tuned decoder to predict molecular properties and explain them in auditable terms.
\bolek{} improves over its Qwen3 base on every task group of a 30-endpoint TDC panel and outperforms the chemistry-specialist TxGemma-9B-Chat on 11 of 17 ADMET endpoints and on four of the five endpoints held out from its training.
Together, these results suggest that targeted modality injection paired with reasoning supervision tied to verifiable features can match domain-specialist systems at a fraction of the parameter count while producing auditable explanations for downstream decisions.

\end{abstract}


\section{Introduction}
\label{sec:introduction}
Machine Learning models have become a central component of modern molecular discovery~\citep{wu2018moleculenet,huang2021tdc,yang2019chemprop,mendezlucio2024mole}, supporting the whole pipeline from hit identification through the late preclinical stage with tasks such as virtual screening, activity \& ADME prediction, toxicity screening, retro-synthesis planning, and lab-in-the-loop optimization cycles.

Yet in high-stakes scientific workflows, a label or a calibrated score is rarely sufficient on its own.
A medicinal chemist deciding whether to synthesise a compound, a pharmacologist triaging a toxicity flag, or a translational scientist building a candidate's profile must understand \emph{why} a molecule is predicted to cross the blood--brain barrier, engage a target, or fail a drug-likeness criterion: downstream decisions hinge not on the prediction alone, but on whether the evidence behind it is mechanistically plausible and chemically meaningful~\citep{jimenezluna2020explainableai}.
When a model returns only a score, the chemist's role collapses into accepting or overriding it; when the model exposes the reasoning behind the score, that reasoning becomes something the chemist can interrogate, correct, and build on, turning prediction into a substrate for design.
The central challenge is therefore not merely to build molecular models that answer accurately, but to build models whose answers are accompanied by explanations that can be inspected, challenged, and connected to chemically meaningful features.

\begin{figure}[t]
\centering \includegraphics[width=\linewidth]{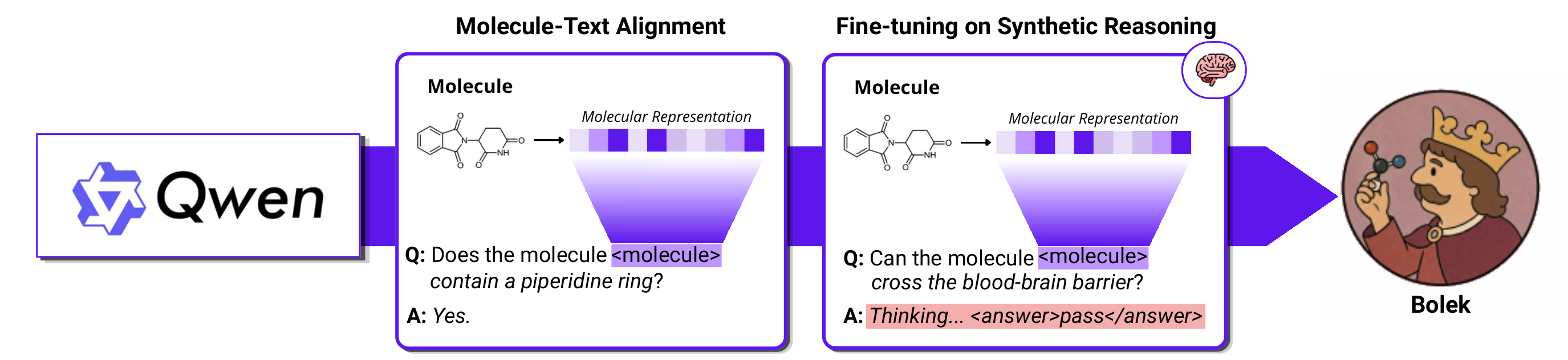} \caption{\bolek{} is built from Qwen3-4B-Instruct in two supervised stages. A projector maps a fixed-size molecular vector into the token-embedding space at the \texttt{<molecule>} position. Alignment teaches the model to answer checkable questions about that molecule; downstream fine-tuning then trains it to predict endpoints, reasoning before it answers.} \label{fig:bolek-graphic}
\end{figure}

The need for grounded explanation is sharpened as molecular prediction moves from standalone benchmark systems into interactive tools used by both human experts and autonomous agents.
In pharmaceutical research, auditable reasoning is necessary for prioritising compounds, diagnosing model failures, and communicating evidence across experimental, computational, and regulatory teams~\citep{jimenezluna2020explainableai}.
At the same time, agentic systems increasingly rely on language models to plan experiments, call tools, and justify decisions~\citep{bran2024chemcrow,boiko2023coscientist,wang2025txgemma}; without grounded reasoning, these systems can produce fluent but chemically unsupported chains-of-thought, errors that then propagate silently through every downstream step.
Moreover, even frontier models are known to occasionally produce contradictory reasoning~\citep{arcuschin2026cotwild}.

Existing approaches tend to address only part of this requirement.
Dedicated molecular predictors built on fingerprints, graph neural networks, and molecular foundation models achieve strong benchmark performance~\citep{rogers2010ecfp,yang2019chemprop,xiong2020attentivefp,xu2019gin,zhou2023unimol,mendezlucio2024mole}, but typically expose only a score, leaving the chemist no way to inspect the evidence behind it.
General-purpose and chemistry-oriented language models can articulate fluent reasoning, but they struggle to read molecular structure from SMILES strings alone~\citep{bartmann2026moleculariq}, and their chains-of-thought routinely cite functional groups, properties, or mechanisms not actually present in the input~\citep{zhang2024chemllmchemicallargelanguage,li2026chemicalqaevaluatingllms,guo2026answerstostates,li2026chemcotscratchpad,yu2024llasmol}.
Tool-calling agents offer a partial remedy by querying external calculators or databases, yet their reasoning remains dependent on orchestration and prompt context rather than on an internal representation that connects molecular evidence to the final decision; recent evaluations show that tool augmentation does not consistently improve over the base LLM on general chemistry tasks, and that the bottleneck lies in the model's own chemical reasoning, not the availability of tools~\citep{yu2025chemtoolagent}.
What high-stakes applications need is competitive prediction paired with explanations that are natively in language and verifiable against the molecule.
Current approaches largely leave this combination unaddressed.

LLM-native multimodal molecular models~\citep{liu2023molca,cao2025instructmol,park2024llamo,le2024molx} appear well placed to close this gap: by attaching a structural encoder directly to a language model, they combine the structural fidelity of dedicated predictors with the explanatory capacity of LLMs and read the molecule natively, without external tools.
Their alignment recipe, however, is largely inherited from image--language pretraining~\citep{radford2021clip,liu2023llava}: the molecular token is trained to map to descriptive prose (PubChem captions, IUPAC names), an objective well suited to naming and describing molecules but not to chemical reasoning, which runs on concrete quantitative evidence (logP, polar surface area, descriptor and substructure counts) rather than qualitative description.
A multimodal LLM that is to reason about molecules, rather than narrate them, needs an alignment objective built around verifiable chemical features from the start.

\paragraph{Contributions.}
Our main contributions are the following.
\begin{itemize}
  \item \textbf{\bolek{}}, a multimodal language model with a deliberately minimal architecture, a molecular embedding (here a Morgan fingerprint~\citep{rogers2010ecfp}) projected onto a single token in an otherwise unmodified Qwen3-4B-Instruct~\citep{yang2025qwen3technicalreport}, which solves downstream endpoints by combining supervision from training labels with literature-derived priors, and generalizes to endpoints excluded from training (\autoref{sec:model-architecture});
  \item A \textbf{suite of alignment tasks}, checkable questions of graded difficulty over more than 850{,}000 molecules, ranging from atoms and bonds, through substructures and computed properties, to descriptions of the whole molecule and of each of its building blocks (\autoref{sec:training-mixture});
  \item \textbf{Synthetic reasoning generation}, a procedure that builds chains-of-thought from literature-derived knowledge, structural decomposition, and the values of task-relevant descriptors, on which the model is then trained to solve downstream tasks (\autoref{sec:downstream-tasks}).
\end{itemize}
We evaluate \bolek{} experimentally and show that (i) trained by supervised fine-tuning alone, it is competitive with a chemistry specialist more than twice its size (\autoref{sec:prediction-quality}), (ii) the recipe transfers across molecular embeddings and backbone sizes, so it is not tied to this representation or this model, and (iii) generating a chain-of-thought before the final prediction helps in some settings, its effect depending on both the endpoint and the molecular representation (\autoref{sec:ablations}).

\section{Related Work}
\label{sec:related-work-tj}
We position our contribution at the intersection of three research lines:
\textbf{text-aligned molecular models} that pair molecular structures with natural-language supervision,
\textbf{chemistry- and therapeutic-specialized LLMs} that operate purely on strings,
and \textbf{supervised molecular property and toxicity prediction}, which defines our evaluation surface.

\paragraph{Text-aligned molecular models.}
Text-aligned molecular work couples structures with natural-language supervision either through shared structure--text spaces or through LLM-native projector bridges.
The first group includes KV-PLM~\citep{zeng2022kvplm}, MoleculeSTM~\citep{liu2023moleculestm}, MoMu~\citep{su2022momu}, and FineMolTex~\citep{li2025finemoltex}; the second includes MolCA~\citep{liu2023molca}, InstructMol~\citep{cao2025instructmol}, LLaMo~\citep{park2024llamo}, 3D-MoLM~\citep{li2024moleculelm}, BioMedGPT~\citep{luo2023biomedgpt}, GIT-Mol~\citep{liu2024gitmol}, MolX~\citep{le2024molx}, MR-MoL~\citep{park2026mrmol}, and EDT-Former~\citep{jing2026edtformer}.
Across both groups, alignment is still mostly \emph{molecule-token to caption} (PubChem descriptions, ChEBI-20 captions~\citep{liu2024chebi20mm}, IUPAC names), with only sporadic substructure or property objectives~\citep{le2024molx,park2024llamo}.
\bolek{} instead pairs a deliberately minimal, fixed Morgan-fingerprint token with feature alignment that asks thousands of checkable questions about molecular structure, descriptors, and substructures.

\paragraph{Chemistry- and therapeutic-specialized LLMs.}
A second line of work adapts general LLMs to chemistry purely through instruction-tuned text.
Galactica~\citep{taylor2022galactica} trains a decoder transformer on a scientific corpus including SMILES;
Mol-Instructions~\citep{fang2024molinstructions} provides a biomolecular instruction corpus that subsequent models build on;
LlaSMol~\citep{yu2024llasmol} fine-tunes Galactica/Llama-2/Code Llama/Mistral on the SMolInstruct corpus.
Tx-LLM~\citep{chaves2024txllm}, fine-tuned from PaLM-2 over $709$ datasets covering $66$ TDC tasks, achieves near- or exceeding-SOTA performance on $43/66$;
its open successor TxGemma~\citep{wang2025txgemma} (Gemma-2; 2B/9B/27B) matches or beats best-in-class on $50/66$ tasks and exceeds specialist models on $26$.
ChemDFM~\citep{zhao2024chemdfm} and BioT5~\citep{pei2023biot5} are further data-centric exemplars.
None of these models, however, exposes the LLM to a \emph{separate} molecular-embedding modality, which is our central design choice.

\paragraph{Molecular property and toxicity prediction.}
Our evaluation lives in the MoleculeNet ecosystem~\citep{wu2018moleculenet}, including Tox21, ToxCast, ClinTox, and SIDER, and the Therapeutics Data Commons (TDC)~\citep{huang2021tdc}, which standardize splits and metrics.
Two baseline families dominate.
Supervised graph and descriptor models include D-MPNN/Chemprop~\citep{yang2019chemprop}, AttentiveFP~\citep{xiong2020attentivefp}, GIN~\citep{xu2019gin}, and Morgan fingerprints~\citep{rogers2010ecfp} paired with random forests or XGBoost.
Self-supervised pretraining methods include ChemBERTa-2~\citep{ahmad2022chemberta2}, MolCLR~\citep{wang2022molclr}, GROVER~\citep{rong2020grover}, Uni-Mol~\citep{zhou2023unimol}, and MolE~\citep{mendezlucio2024mole}.

\paragraph{Positioning.}
\bolek{} occupies the gap left by these lines: it replaces molecule-as-caption alignment with atomic questions about structure, descriptors, and substructures, posed against a parameter-free Morgan-fingerprint interface~\citep{rogers2010ecfp}.
Prediction quality competitive with chemistry-specialized LLMs such as TxGemma~\citep{wang2025txgemma} then emerges as a consequence of alignment rather than a separate design target.

\section{Method}
\label{sec:method}
\bolek{} combines three components, shown in \autoref{fig:bolek-graphic}.
The \textbf{molecular embedding} is projected onto a single token in the token-embedding space, so the model reads the structure natively rather than from a SMILES string.
The \textbf{suite of alignment tasks} (\autoref{fig:alignment-dataset}) teaches it to answer checkable questions about that token, ascending from atoms and bonds through substructures and computed properties to whole molecules and their parts.
\textbf{Synthetic reasoning generation} (\autoref{fig:downstream-datasets}) then ties those facts to endpoint labels, grounding each rationale in a literature-derived mechanism and in concrete RDKit descriptors relevant for the endpoint.
The architecture is minimal and universal, admitting any fixed-size molecular embedding and any base language model, so the two protocols carry over unchanged to other representations and backbones.

\subsection{Molecular Embedding}
\label{sec:model-architecture}

\bolek{} starts from Qwen3-4B-Instruct~\citep{yang2025qwen3technicalreport} and adds a two-layer projector that turns a fixed-size molecular vector into one additional token at the \texttt{<molecule>} position, described in Appendix~\ref{app:projector}.
Everything else is embedded and decoded as usual, so the language model is otherwise unchanged.
We instantiate the vector with 2048-bit, radius=2 Morgan fingerprints~\citep{rogers2010ecfp}, which are well established in cheminformatics and carry strong predictive signal for the QSAR-style tasks considered here.

Nothing in the design depends on that choice, so the representation is a swappable input.
\autoref{tab:ablations} compares Morgan fingerprints with Uni-Mol~\citep{zhou2023unimol} and Molformer-XL~\citep{ross2022molformer} pooled embeddings of 512 and 768 dimensions, and with the canonical SMILES string~\citep{weininger1988smiles} given in the prompt.
SMILES is the control, removing the projector entirely.
All four follow the same two-stage training procedure, so differences between them primarily reflect the input representation rather than the instruction-tuning recipe.

\subsection{Alignment Tasks}
\label{sec:training-mixture}

\begin{figure}[t]
\centering
\includegraphics[width=\linewidth]{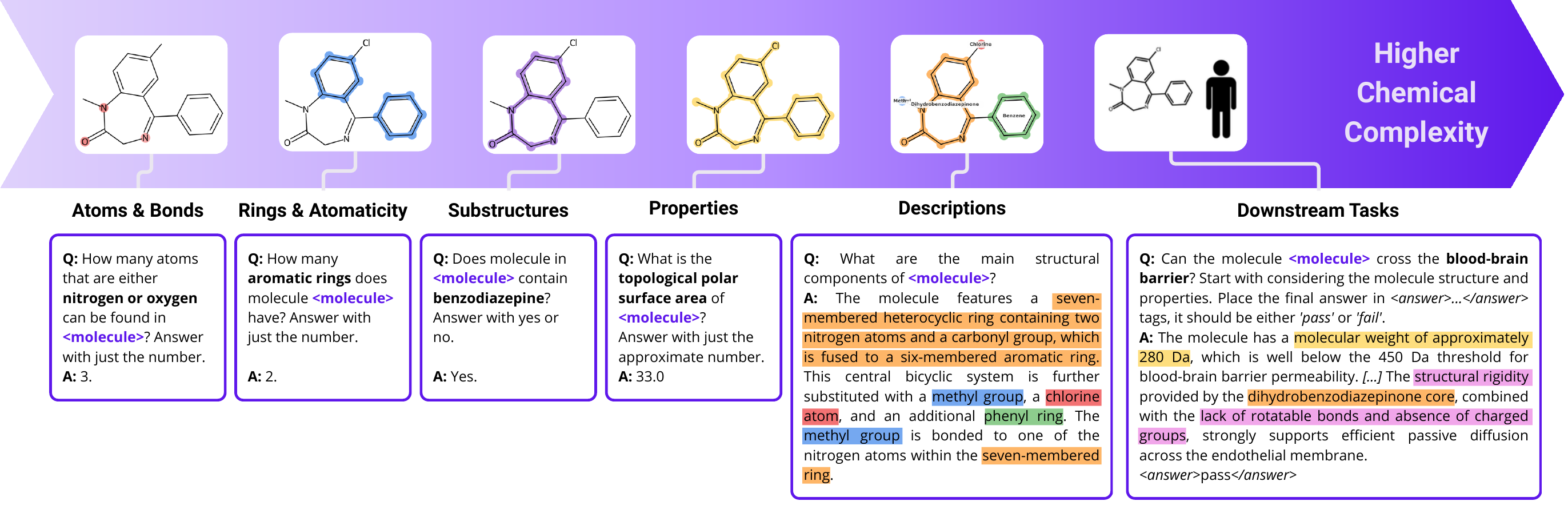}
\caption{The alignment mixture asks checkable questions about one molecule, ordered by the chemical complexity of the evidence each answer needs.
Structured questions have a single verifiable answer; the description and downstream tasks are free text.}
\label{fig:alignment-dataset}
\end{figure}

The alignment tasks teach the model to read the molecule token, climbing a ladder of chemical abstraction (see \autoref{fig:alignment-dataset}), from atoms and bonds, through substructures and computed properties, to descriptions of the whole molecule and of its named building blocks in terms of local physicochemical properties such as lipophilicity, polarity, and protonation.
The mixture draws on more than 850{,}000 molecules from public corpora described in Appendix~\ref{app:alignment-data}, asking the model to classify 1{,}403 substructure patterns and regress 88 RDKit or Mordred descriptors~\citep{rdkit2024,moriwaki2018mordred} alongside the whole-molecule and per-part descriptions.
The free-text molecule descriptions are therefore grounded in verifiable chemical facts, unlike the image captions used in vision--language alignment.

\subsection{Synthetic Reasoning Generation}
\label{sec:downstream-tasks}

Downstream supervision then teaches the model to predict what the molecule does in an assay or an organism.
The suite has 30 TDC binary classification tasks~\citep{huang2021tdc}, of which 25 are used for training and 5 are held out for evaluation only.
The 25 comprise 17 ADMET and bioactivity endpoints, spanning mutagenicity, blood--brain-barrier penetration, bioavailability, intestinal absorption, hERG liability, HIV activity, CYP inhibition and substrate status, clinical toxicity, skin reaction, muscarinic M1 agonism, and SARS-CoV-2 activity, together with 8 Tox21 nuclear-receptor and stress-response assays.
Each endpoint is trained in two answer formats, a direct yes/no format that asks only for the binary label, and a chain-of-thought (CoT) format that reasons over structure and properties before emitting the label in \texttt{<answer>...</answer>} tags (see representative prompts in Appendix~\ref{app:training-task-examples}).
Training labels are balanced by upsampling the minority class, while validation and test labels follow the TDC scaffold splits unchanged.

The held-out endpoints are chosen one per task family, each for a specified relationship to the trained panel, so that a transfer result can be interpreted and not only measured.
They form a ramp of transfer distance.
M1 antagonist inverts the polarity of the trained M1 agonist task, CYP2D6 inhibition shares the Veith assay with the trained CYP inhibition tasks and its enzyme with a trained CYP substrate task but belongs to a different gene subfamily with its own pharmacophore, PAMPA NCATS permeability moves from the trained human absorption endpoints to an in vitro assay format, Tox21 PPAR-\(\gamma\) is a lipid-sensing receptor unlike the steroid-binding receptors among the trained Tox21 assays, and Pgp Broccatelli has no trained relative and requires transfer across task families.

\begin{figure}[t]
\centering
\includegraphics[width=0.92\linewidth]{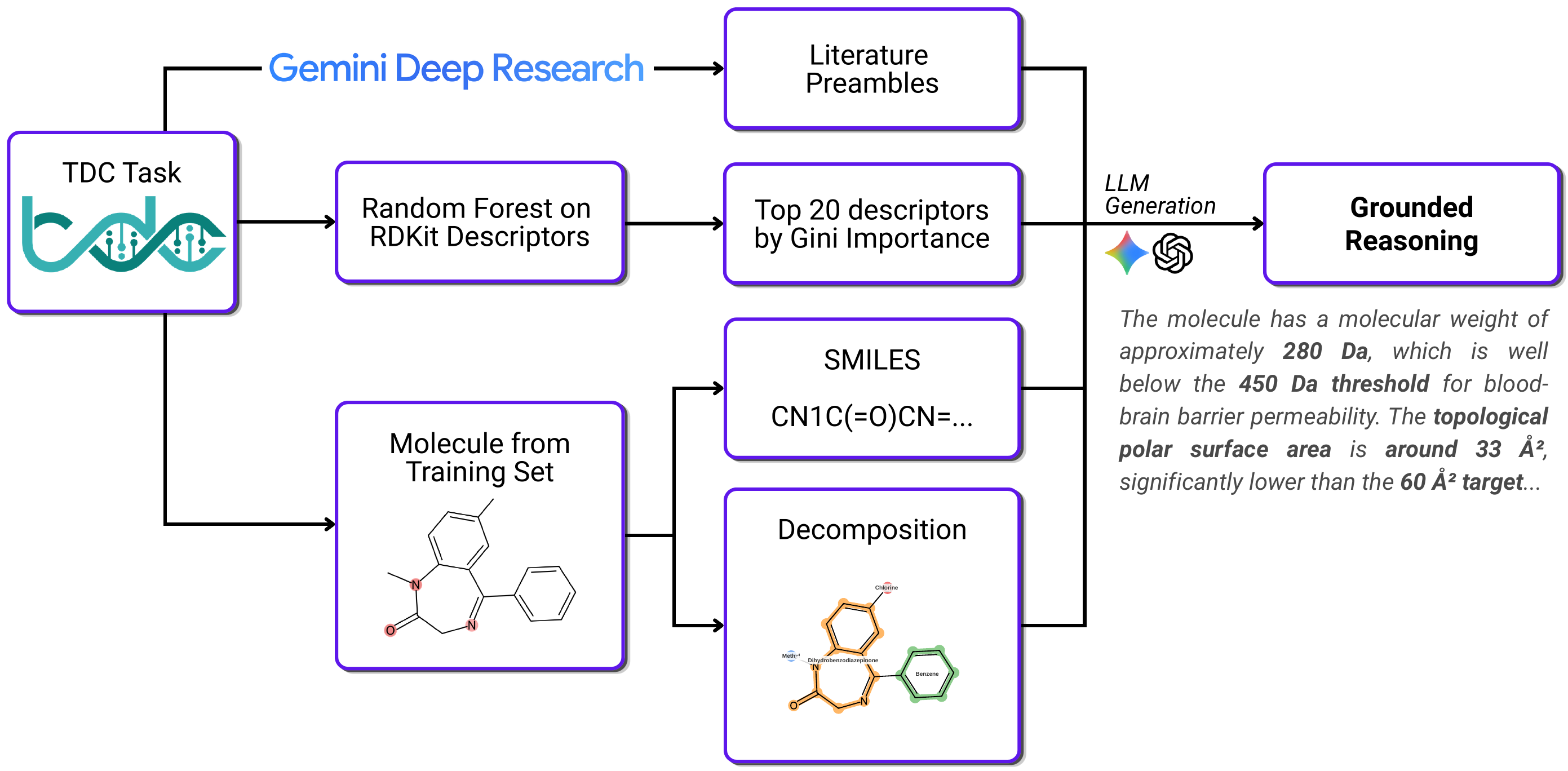}
\caption{Building a chain-of-thought target for one endpoint.
The teacher writes the rationale offline, from inputs \bolek{} itself never sees at training time.}
\label{fig:downstream-datasets}
\end{figure}

\paragraph{Descriptor-grounded reasonings.}
The chain-of-thought targets are synthetic but chemically constrained.
A teacher model receives the molecule's SMILES and its bond-cutting decomposition, the values of task-relevant descriptors, and assay context (see \autoref{fig:downstream-datasets}).
The assay context is a \emph{literature preamble} for the ADMET endpoints and an \emph{assay card} for the Tox21 assays, both mined from the literature as Appendix~\ref{app:method-details} describes.
The descriptors are the top 20 ranked by a random forest fit on that task's own training split, so the rationale is steered towards the features that carry signal for that endpoint.
Tox21 rationales are additionally structured as an adverse-outcome pathway.
A rationale is kept only when its tagged final answer matches the ground-truth label, and molecules that never yield one are excluded from chain-of-thought training but still appear in yes/no training.
The teacher alone sees these inputs, and \bolek{} is trained by teacher forcing on the generated rationale, receiving only the downstream question and the molecule token.
The high-throughput screens (Tox21, HIV, M1, SARS-CoV-2) are large and dominated by inactives, so generating a rationale for every molecule would let them swamp the downstream mixture.
We therefore cap how many inactives accompany each active, selecting them for structural relevance on the Tox21 assays, with the full selection rules, teacher models, and attempt budgets given in Appendix~\ref{app:cot-construction}.

\paragraph{Literature-mined reasonings.}
A second source of chain-of-thought targets is mined from published literature rather than derived from descriptors, one molecule and one endpoint at a time.
An agent searches PubMed and returns a report whose every step cites a source and carries an evidence level, which a second call condenses into a short rationale running from structure to conclusion (Appendix~\ref{app:mined-reasoning}).
Mining is restricted to molecules matched to ChEMBL compounds, since anonymous screening compounds carry no literature, and yields 2{,}214 rationales over 13 ADMET endpoints, about 9\% of the chain-of-thought draws on those endpoints.

\subsection{Training}
\label{sec:training}

Training runs in two supervised stages, 8{,}250 steps on the alignment tasks alone, then 1{,}750 steps on downstream supervision alone, in both cases training the full backbone and the projector with the same next-token objective over the answer (Appendix~\ref{app:training-config}).
Every downstream training prompt also carries the assay context, injected with 50\% dropout, so the model learns to answer both with and without assay context, and remains usable when none is available.
\autoref{tab:ablations} compares this recipe against training alignment and downstream supervision together in one stage, and against continuing it with reinforcement learning on answer correctness (Appendix~\ref{app:rl}).
Appendix~\ref{app:size-ablation} compares backbones from 0.6B to 14B parameters under the same recipe, where downstream gains plateau from 4B while training cost grows roughly linearly, motivating 4B as the main configuration.

\section{Experiments}
\label{sec:experiments}
\subsection{Evaluation}
\label{sec:evaluation-setup}

\begin{table}[t]
\centering
\caption{Mean ROC AUC per task group.
An underlined cell differs from \bolek{} CoT by more than a 95\% paired bootstrap interval over the shared test molecules; a plain cell does not.
YN = yes/no answer, CoT = chain-of-thought.}
\label{tab:downstream}
\small
\begin{tabular}{llccccc}
\toprule
Model & Trained on & Format & ADMET (17) & Tox21 (8) & Held-out (5) \\
\midrule
Qwen3-4B-Instruct & none & YN & \underline{0.559} & \underline{0.520} & \underline{0.588} \\
Qwen3-4B-Instruct & none & CoT & \underline{0.653} & \underline{0.557} & \underline{0.690} \\
GPT-6-Sol\(^{\dagger\S}\) & none & YN & \underline{0.687} & \underline{0.576} & \underline{0.690} \\
GPT-6-Sol\(^\dagger\) & none & CoT & \underline{0.730} & \underline{0.607} & \underline{0.714} \\
\midrule
\bolek{} & ADMET + Tox21 & YN & \underline{0.743} & 0.747 & \underline{0.770} \\
\bolek{} & ADMET + Tox21 & CoT & 0.760 & 0.762 & 0.799 \\
\midrule
TxGemma-9B-Chat & all & -- & \underline{0.714} & \underline{0.792} & \underline{0.738} \\
TxGemma-9B-Predict & all & -- & \underline{0.800} & \underline{0.899} & \underline{0.855} \\
Morgan RF & all & -- & 0.763 & \underline{0.737} & 0.812 \\
Descriptor RF & all & -- & 0.780 & \underline{0.787} & 0.822 \\
Uni-Mol & all & -- & 0.746 & \underline{0.732} & \underline{0.757} \\
Molformer-XL & all & -- & 0.754 & \underline{0.722} & 0.791 \\
Chemprop D-MPNN & all & -- & 0.746 & 0.748 & 0.824 \\
\bottomrule
\end{tabular}

\vspace{2pt}
{\footnotesize \(^\dagger\)Scored from 5 rollouts per molecule rather than 50.
\(^\S\)Averaged over 16 of the 17 ADMET endpoints: the API refused SARS-CoV-2 Touret in this format.}
\end{table}

We evaluate downstream molecular endpoints with ROC AUC for classification and Spearman correlation plus mean absolute error for regression~\citep{huang2021tdc}.
All downstream metrics use the corresponding TDC scaffold-split test set unless otherwise noted.
For \bolek{} and Qwen3, AUC scores are estimated from repeated stochastic generations by parsing final labels, while TxGemma-9B-Chat scores use the answer-choice token logit ratio with the TDC prompts provided in the model release.
Unless stated otherwise, evaluation supplies the same assay context the model was trained with, and \bolek{} and Qwen3 receive it alike, with \autoref{tab:ablations} ablating it on both the training and the inference side.
Appendices~\ref{app:method-details} and~\ref{app:txgemma-comparison} give the full scoring, rollout, and baseline details, and the TxGemma-9B-Chat protocol.

\subsection{Prediction Quality}
\label{sec:prediction-quality}

\autoref{tab:downstream} places \bolek{} against other models, grouped by their respective training panel.

\bolek{} gains 0.11 on ADMET, 0.21 on Tox21 and 0.11 on held-out over Qwen3-4B-Instruct answering in the same format with the same assay knowledge, significant in every group.
GPT-6-Sol, a frontier model given the same molecular context, falls a significant 0.030 behind \bolek{} on ADMET, 0.16 on Tox21 and 0.09 on held-out.

Against TxGemma-9B-Chat, \bolek{} is ahead by a significant 0.046 on ADMET and by 0.061 on the five endpoints that are held out from \bolek{} but not from TxGemma, and behind by 0.030 on Tox21.

A similar split appears against the supervised specialists.
On the endpoints \bolek{} trains on it holds its own: on ADMET it is level with all of them, and on Tox21 it is ahead of Morgan RF, Uni-Mol and Molformer-XL, and behind only Descriptor RF.
On the held-out five none of them beats it and Uni-Mol falls behind, though each was refitted on those endpoints' own train splits.
Finally, TxGemma-9B-Predict is ahead everywhere. It is the same backbone without a conversational objective, so that gap estimates what \bolek{} gives up producing an explanation rather than a score.

Three TDC regression endpoints, none of them trained on, test whether the same interface transfers to a different answer type (Appendix~\ref{app:heldout-regression}).
\bolek{} is the only model that ranks plasma protein binding at all, 0.340 Spearman against 0.001 for Qwen3 and \(-0.001\) for TxGemma-9B-Chat, while TxGemma leads on lipophilicity and solubility.

The two answer formats differ by endpoint rather than uniformly~\citep[cf.][]{sprague2025cotornot}, as \autoref{fig:cot-gain} in Appendix~\ref{app:cot-gain} resolves per endpoint.
Chain-of-thought is at least as strong as yes/no in every task group, and it gains most exactly where the model has never seen the endpoint: 0.029 on the held-out five, against 0.017 on ADMET and 0.015 on Tox21.
Its significant losses concentrate on the Veith CYP-inhibition tasks, where abundant labels let direct training learn finer substructure evidence.

\subsection{Ablations}
\label{sec:ablations}

\begin{table}[t]
\centering
\caption{Ablations of \bolek{}, reported as the mean ROC AUC per task group as in \autoref{tab:downstream}.
Each block changes one design choice and repeats \bolek{} as its reference row.
Variants trained without chain-of-thought are not evaluated on it (--).
An underlined cell differs from that reference by more than a 95\% paired bootstrap interval over the shared test molecules.
All variants are evaluated under the default protocol, except the last block, which varies that protocol on the main model.}
\label{tab:ablations}
\footnotesize
\begin{tabular}{lcccccc}
\toprule
& \multicolumn{2}{c}{ADMET (17)} & \multicolumn{2}{c}{Tox21 (8)} & \multicolumn{2}{c}{Held-out (5)} \\
\cmidrule(lr){2-3}\cmidrule(lr){4-5}\cmidrule(lr){6-7}
Variant & CoT & YN & CoT & YN & CoT & YN \\
\midrule
\multicolumn{7}{l}{\emph{Molecular representation}} \\
Morgan fingerprint (\bolek{}) & 0.760 & 0.743 & 0.762 & 0.747 & 0.799 & 0.770 \\
Uni-Mol & \underline{0.743} & 0.733 & \underline{0.747} & 0.751 & \underline{0.769} & \underline{0.742} \\
Molformer-XL & 0.753 & 0.747 & 0.770 & 0.761 & \underline{0.775} & 0.776 \\
SMILES & 0.746 & 0.749 & \underline{0.742} & 0.760 & \underline{0.777} & \underline{0.736} \\
\midrule
\multicolumn{7}{l}{\emph{Alignment stage and supervision}} \\
Alignment, then CoT and YN (\bolek{}) & 0.760 & 0.743 & 0.762 & 0.747 & 0.799 & 0.770 \\
Alignment, then YN only & -- & \underline{0.716} & -- & \underline{0.646} & -- & \underline{0.640} \\
No alignment, CoT and YN & 0.748 & 0.726 & 0.766 & \underline{0.729} & 0.788 & \underline{0.720} \\
No alignment, YN only & -- & \underline{0.683} & -- & \underline{0.634} & -- & \underline{0.636} \\
No alignment, projector only & \underline{0.654} & \underline{0.650} & \underline{0.668} & \underline{0.684} & \underline{0.660} & \underline{0.639} \\
\midrule
\multicolumn{7}{l}{\emph{Training pipeline}} \\
Alignment then downstream (\bolek{}) & 0.760 & 0.743 & 0.762 & 0.747 & 0.799 & 0.770 \\
Alignment then downstream, then GRPO RL & 0.759 & 0.742 & \underline{0.769} & \underline{0.724} & 0.792 & 0.765 \\
Single joint stage & 0.755 & 0.755 & \underline{0.770} & \underline{0.773} & 0.793 & 0.784 \\
\midrule
\multicolumn{7}{l}{\emph{In-context assay knowledge during training}} \\
Injected with 50\% dropout (\bolek{}) & 0.760 & 0.743 & 0.762 & 0.747 & 0.799 & 0.770 \\
Never injected & \underline{0.736} & \underline{0.718} & \underline{0.719} & 0.741 & 0.795 & \underline{0.794} \\
\midrule
\multicolumn{7}{l}{\emph{In-context assay knowledge at inference}} \\
In the prompt (\bolek{}) & 0.760 & 0.743 & 0.762 & 0.747 & 0.799 & 0.770 \\
Absent from the prompt & 0.756 & 0.750 & 0.763 & 0.744 & 0.784 & \underline{0.744} \\
\bottomrule
\end{tabular}
\end{table}

The representation variants each repeat the alignment stage against their own encoder, so every downstream stage begins from a checkpoint aligned to its representation.

\paragraph{Molecular representation.}
No representation separates from the Morgan fingerprint in its own favour on any task group.
Uni-Mol falls behind it in four cells and SMILES in three, both held-out cells among them in each case, and Molformer-XL trails only on held-out chain-of-thought.
Where the learned encoders lead, on Tox21, they lead by 0.014 or less and no interval excludes zero.
The fingerprint is also the cheapest of the four, computed without a learned encoder, and it keeps the substructure vocabulary of the alignment stage interpretable.

\paragraph{Alignment stage and supervision.}
The alignment stage and chain-of-thought supervision both raise downstream ROC AUC, and their gains add up: \bolek{}, trained with both, scores highest in every yes/no column of the block, and no variant scores significantly above it in any column.
Chain-of-thought supervision is the larger of the two; removing the chain-of-thought traces from downstream training lowers ROC AUC most on Tox21 and on the held-out five, with or without the alignment stage.
The alignment stage shows its effect most clearly when downstream training uses yes/no questions only: the aligned model then scores higher than a model trained from Qwen3-4B without alignment on the ADMET and Tox21 endpoints seen during training, significantly so on ADMET.
When the chain-of-thought traces are present, removing the alignment stage lowers ROC AUC less, because the traces restate alignment-stage properties such as polar surface area.
Training only the projector scores significantly lower than \bolek{} in every column, so the language model itself has to adapt to the molecular input.
Appendix~\ref{app:alignment-ablation} resolves this block per task.

\paragraph{Training pipeline.}
Continuing the supervised model with GRPO reinforcement learning~\citep{shao2024deepseekmath} trades one Tox21 format for the other, 0.022 off yes/no and 0.007 onto chain-of-thought, and changes nothing on ADMET or the held-out five.
Training both mixtures together in a single stage is ahead on Tox21 in both formats, by 0.008 on chain-of-thought and 0.027 on yes/no, and is level everywhere else.
We report the two-stage model because it leads on chain-of-thought for ADMET and for the held-out five, by 0.005 to 0.006 without separating, and treat the single-stage recipe as an open question rather than a settled one.

\paragraph{Assay knowledge during training.}
Injecting cards and preambles during training, with 50\% dropout, is what makes the model able to use them at all.
It is worth 0.025 on ADMET chain-of-thought, 0.025 on ADMET yes/no and 0.042 on Tox21 chain-of-thought.
The gain does not reach every group: the never-injected variant is ahead by 0.024 on held-out yes/no, and the remaining cells are level.

\paragraph{Assay knowledge at inference.}
Supplying the same knowledge at inference matters most where the model has no trained-in knowledge of the endpoint.
It is worth 0.026 on held-out yes/no, the only cell that separates, against 0.014 on held-out chain-of-thought and nothing on any trained group.
This is the expected shape if the preamble supplies assay context that training already provides for the trained endpoints.

\section{Discussion and Limitations}
\label{sec:discussion}
\paragraph{Molecular alignment exposes reusable signal.}
The main result of this study is that molecular alignment can turn a general instruction-tuned language model into a stronger molecular predictor while preserving its ability to produce natural-language rationales.
\bolek{} improves substantially over Qwen3 on downstream binary classification, transfers to held-out endpoints, and remains competitive with stronger chemistry-oriented baselines on several tasks.
Some endpoints improved more than others.
This may reflect how much can be predicted from molecular structure and descriptors alone, and how much depends on assay conditions or biological factors that the model does not observe.

\paragraph{Reasoning supervision is still imitation rather than verification.}
The chain-of-thought stage teaches \bolek{} to organize molecular evidence into task-specific reasoning, but it does not directly optimize the truth of each intermediate statement.
That truth matters, because the answer follows the rationale.
When its rationale is replaced by one written for another molecule of the same endpoint, with the verdict removed, \bolek{} gives the label that rationale argues for over 98\% of test molecules (Appendix~\ref{app:counterfactual}).
This limitation is visible in the mixed effect of chain-of-thought supervision: it helps on some small-data pharmacophore-like tasks, but can hurt when the supervised rationale prior is coarser than the signal available in the training labels.
Because CoT targets are kept only when the teacher reaches the ground-truth label, the CoT subset can also be biased toward examples that are easier for the teacher, making more complete and less selective CoT generation an important direction for improving reasoning supervision.
Consistent with this, continuing the supervised model with GRPO on answer correctness alone trades one answer format for the other rather than improving either overall (\autoref{tab:ablations}), which suggests the useful reward signal is not the final label.
Rewards that check the rationale itself are a natural next step, and in chemistry much of a rationale can be checked exactly~\citep{guo2026answerstostates}: rewarding a correct answer only when its rationale contains no fabricated molecular claims reduces fabrication without costing accuracy~\citep{li2026chemcotscratchpad}.
Imitation nonetheless carries further than the endpoints it was written for: the gain from reasoning is largest on the five held-out endpoints (\autoref{fig:cot-gain}), where the model has no endpoint-specific training signal to fall back on.

\paragraph{The molecular projector is intentionally minimal.}
\bolek{} uses a single projected Morgan fingerprint token, which makes the architecture simple and the contribution easy to isolate.
However, Morgan fingerprints~\citep{rogers2010ecfp} are also a bottleneck: they compress molecular structure into hashed local environments and discard information that may matter for stereochemistry, conformation, long-range geometry, target binding, quantum effects, and assay-specific context.
The same interface accepts richer fixed-size embeddings, and \autoref{tab:ablations} already swaps in two learned ones: on the task groups measured so far a Molformer-XL embedding is ahead of the fingerprint, so the representation rather than the interface looks like the binding constraint.
Multi-token representations~\citep{jing2026edtformer}, protein-conditioned representations, and ensembles that combine several molecular views remain open~\citep{li2024moleculelm,liu2023molca,park2024llamo,mendezlucio2024mole}.

\paragraph{Native multimodality and tool use are complementary.}
Tool-calling systems can compute exact descriptors, retrieve external facts, and verify specific claims~\citep{bran2024chemcrow,boiko2023coscientist}, while \bolek{} places compact molecular evidence directly inside the language model.
The broader point is that competitive prediction and auditable reasoning need not trade off against each other.
\bolek{} is one step toward that combination: it improves prediction, exposes reusable molecular signal on held-out endpoints, and answers in a form a chemist can interrogate rather than simply accept.

\section{Conclusion}
\label{sec:conclusion}
\bolek{} pairs a molecular embedding projected onto a single token with a suite of alignment tasks that ground that token in checkable chemical facts, and with synthetic reasoning generation that supervises its chains-of-thought.
Alignment allows the model to apply its literature priors to a molecule it reads directly, and the results show \bolek{} combining both sources of knowledge, the training labels and those priors, and generalizing to endpoints excluded from training.
With a 4B backbone it achieves performance comparable to the evaluated TxGemma-9B-Chat, with higher mean scores on the ADMET endpoints, similar results on the five held-out endpoints, and lower scores on Tox21.

The benefit of answering in chain-of-thought rather than yes/no depends on the endpoint.
It is largest on the five endpoints the model never trained on.
One possible explanation is that a rationale lets the model fall back on general chemical reasoning where it has no endpoint-specific training signal to draw on.
The ablation results are consistent with this interpretation.

We also observed smaller chain-of-thought gains on larger datasets.
The clearest losses occurred on four of the five Veith CYP-inhibition tasks, while HIV improved despite being one of the largest datasets.
This pattern suggests that the amount of available supervision may influence the benefit of a rationale.
In our panel, size varies together with assay type and with the rationale-generation procedure, so its contribution cannot be cleanly separated.

These results show that a simple molecular projector and supervised training can support competitive prediction in a compact language model.
They also show that generating a rationale can help some tasks and hurt others.
Understanding these differences, and evaluating the chemical correctness of the rationales themselves, are the next steps.

\subsection*{AI Use Statement}
In this work, we used generative AI tools to generate synthetic data sets, to clean and reformat data, to implement methods, and to design and give feedback on experiments.
The first three are components of the method rather than aids to reporting it: the synthetic chain-of-thought targets of \autoref{sec:downstream-tasks}, the per-part decomposition supervision of the alignment mixture, and the literature preambles and assay cards are all model-generated, by the systems named in Appendix~\ref{app:method-details} and listed with their terms of use in \autoref{tab:external-asset-licenses}.
Generative AI additionally assisted in writing the training, evaluation and analysis code, in choosing and reviewing analyses such as the paired bootstrap procedure of \autoref{sec:evaluation-setup}, and in summarising differences across large sets of per-endpoint results.
We have not used generative AI tools to interpret those results, which the authors did themselves, nor to propose or refine hypotheses, to develop theoretical models or conceptual frameworks, to formulate mathematical claims, to provide ingredients for or write proofs, to assist with translation, or to support qualitative and thematic data analysis; the remaining tasks requiring disclosure are not applicable to this work.

Additionally, we used generative AI tools to create and modify scientific figures, to create and edit software code, to draft and edit parts of this paper for readability, to search for and summarise existing literature, and to format references.

We have reviewed all AI-assisted work.
Every AI-assisted change was directed by an author and reviewed by at least one other author before it entered the repository, under a commit policy that records the level of AI involvement on each commit.
All numbers reported in this paper are regenerated from stored model outputs by the scripts released with it rather than written into the text by a model, and the script that produces each table is named in its caption.
We take responsibility for the final content of this work, including text, claims or artifacts produced with the aid of generative AI.

\bibliographystyle{unsrtnat}
\bibliography{references}

\appendix

\section{Method}
\label{app:method-details}
\subsection{Evaluation Protocol}

We quantify test-set sampling uncertainty with a 95\% percentile bootstrap over test molecules, using 2{,}000 resamples~\citep{efron1994bootstrap}.
Classification resamples are stratified by label, drawing positives and negatives to their original counts, while regression resamples are unstratified.
Point estimates are always computed on the full test set, and per-molecule scores are held fixed across resamples, so the intervals exclude generation noise, which Appendix~\ref{app:rollout-sensitivity} characterizes separately.
Model comparisons are paired, scoring both models on the same resamples and restricting to the molecules they share, and a difference is called significant when its 95\% interval excludes zero.
Means over tasks average the per-task bootstrap distributions, so their intervals reflect molecule sampling within a fixed set of endpoints.
\label{app:evaluation-protocol}

Alignment binary tasks are evaluated by accuracy, and alignment regression tasks by Pearson correlation.
Downstream classification tasks use ROC AUC, and downstream regression tasks use Spearman correlation and mean absolute error.
Unless stated otherwise, all downstream metrics are computed on the corresponding TDC scaffold-split test set.

For TxGemma-9B-Chat, we use the TDC prompts provided in the model release and compute classification scores from the answer-choice token logit ratio.
For \bolek{} and Qwen3, CoT AUC scores are estimated from repeated stochastic generations because the final-answer probability is conditioned on the generated rationale and can saturate after the model has committed to an answer.
This rollout fraction is a Monte Carlo estimate of the model's positive-answer probability, analogous in the limit to direct probability or logit scoring when those scores are accessible.
We run 50 generations per molecule at temperature \(0.6\), parse the final label, and use the fraction of parseable positive labels as the score.
For regression endpoints, we parse the numeric prediction from the generated answer and compare it with the ground-truth value in the original TDC units.
We use a 300-tree random forest trained on RDKit descriptors~\citep{rdkit2024} to rank per-task descriptors for CoT construction.

\subsection{Projector}
\label{app:projector}

A two-layer projector \(g_{\phi}\), with SiLU activation~\citep{elfwing2018silu} and hidden width twice the Qwen3 embedding dimension, maps the molecular vector \(\mathbf{m}\) into the token-embedding space:
\[
\begin{aligned}
g_{\phi}(\mathbf{m}) &= W_2 \operatorname{SiLU}(W_1 \mathbf{m} + b_1) + b_2, \\
\mathbf{e}_i &=
\begin{cases}
g_{\phi}(\mathbf{m}) & \text{if } x_i = \text{\texttt{<molecule>}}, \\
E(x_i) & \text{otherwise}.
\end{cases}
\end{aligned}
\]
The mixed sequence is then processed by the standard causal decoder, leaving the language-model architecture otherwise unchanged.

\subsection{Training Configuration}
\label{app:training-config}

All variants are optimized with the same autoregressive instruction-tuning objective in both stages.
Prompt tokens are masked from the loss and supervision is applied only to assistant responses, so alignment tasks, direct yes/no downstream tasks, and chain-of-thought downstream tasks are all trained as next-token prediction over the answer.
We train the full Qwen3 backbone in every variant except the projector-only ablation of \autoref{tab:ablations}, and the projector alongside it wherever the molecule enters as a vector.
The downstream stage does not replay alignment data.

\begin{table}[h]
\centering
\caption{Training configuration for the two supervised stages of \bolek{}.
The ablation variants of \autoref{tab:ablations} keep these settings and change only the aspect under test.}
\label{tab:training-config}
\small
\begin{tabular}{lcc}
\toprule
& Alignment & Downstream \\
\midrule
Initialized from & Qwen3-4B-Instruct & alignment checkpoint \\
Steps & 8{,}250 & 1{,}750 \\
Maximum sequence length & 256 & 1{,}280 \\
Effective batch size & 256 & 256 \\
Backbone learning rate & \texttt{5e-6} & \texttt{5e-6} \\
Projector learning rate & \texttt{5e-5} & \texttt{5e-5} \\
Schedule & \multicolumn{2}{c}{linear decay, no warmup, zero weight decay} \\
Precision & \multicolumn{2}{c}{bf16 fully sharded data parallelism} \\
Hardware & 4\(\times\)H100 & 8\(\times\)H100 \\
Wall time & \(\approx\)5\,h & \(\approx\)2.5\,h \\
\bottomrule
\end{tabular}
\end{table}

The longer downstream sequence length accommodates chain-of-thought targets, which the yes/no format does not need.

\subsection{Reinforcement Learning}
\label{app:rl}

The reinforcement-learning variant of \autoref{tab:ablations} continues the supervised model rather than replacing it.
It starts from the downstream checkpoint and runs 2{,}000 steps of group-relative policy optimization~\citep{shao2024deepseekmath} with the DAPO loss~\citep{yu2025dapo}, as implemented in TRL, generating with vLLM.
Only the chain-of-thought targets are optimized; the yes/no format is never trained, which is worth keeping in mind when reading that variant's yes/no column.
Prompts are balanced by label within each endpoint and equalized across the 25 of them, capped by the smallest available label pool, so neither a class nor an endpoint dominates a batch.

The reward is dominated by answer correctness.
A completion scores 1 when the label inside its \texttt{<answer>} tags matches the ground truth and 0 otherwise, at weight 1.0.
A second term scores 0.5 for a parseable \texttt{<answer>} tag, at weight 0.1, which keeps the output format from drifting.
A third term, which would penalize a completion that opens with \texttt{<answer>} and so skips the rationale entirely, carries weight 0 in this run.
No term inspects the content of the rationale, so this variant tests answer correctness alone.

\begin{table}[h]
\centering
\caption{Reinforcement-learning configuration for the ablation variant of \autoref{tab:ablations}.}
\label{tab:rl-config}
\small
\begin{tabular}{ll}
\toprule
Setting & Value \\
\midrule
Initialized from & downstream checkpoint, step 1{,}750 \\
Steps & 2{,}000 \\
Generations per prompt & 8 \\
Generation batch size & 256 \\
Maximum completion length & 512 \\
Sampling temperature & 0.7 \\
Learning rate & \texttt{5e-7} \\
KL coefficient & 0.01 \\
Effective batch size & 64 \\
Gradient-norm clipping & 1.0 \\
Reward weights (accuracy / format / rationale) & 1.0 / 0.1 / 0.0 \\
\bottomrule
\end{tabular}
\end{table}

\section{Data}
\label{app:data}
\subsection{Alignment Data}
\label{app:alignment-data}

Alignment exposes the molecule to the model through natural-language question--answer exchanges.
The molecule pool combines four sources: 700{,}000 sampled molecules from MolPILE~\citep{adamczyk2025molpile}, 90{,}000 KnowMol molecules with multi-level structural and property annotations~\citep{yang2025knowmol}, 26{,}402 ChEBI-20-MM molecules with natural-language descriptions~\citep{liu2024chebi20mm}, and a name-prediction set with 32{,}936 molecules from an RCSB PDB-derived curated docking library~\citep{Berman2000} plus 796 ZINC20 molecules~\citep{sterling2020zinc20}.
Each alignment task uses 5--20 paraphrased templates; representative examples are shown in Appendix~\ref{app:training-task-examples}.

Free-text tasks include whole-molecule descriptions, aspect-specific descriptions of decomposed molecular parts, molecule naming, and SMILES recovery from the fingerprint.
For the eight decomposition tasks, each molecule is decomposed into named structural parts, each part is annotated with local properties such as logP, hydrogen-bond donors and acceptors, stereochemistry, polarity, protonation, and partial charge, and Gemini~2.5~Flash generates the supervised answer from this annotated decomposition.

Substructure classification asks whether a molecule contains a named pattern.
The vocabulary contains 534 simple atom-and-bond patterns, 153 MACCS keys, 73 RDKit fragment descriptors~\citep{rdkit2024}, and 643 textbook or alert patterns assembled from RDKit functional-group definitions, RDKit hierarchical filters, SureChEMBL structural alerts~\citep{papadatos2016surechembl}, and SMARTS-RX~\citep{kogej2025smartsrx}.
Items are partitioned into seen and unseen subsets, and labels are balanced through targeted molecule sampling.
Descriptor prediction covers 88 RDKit and Mordred descriptors~\citep{moriwaki2018mordred}, including atom and bond counts, ring counts, molecular weight, and logP.

\subsection{Molecule Decomposition}
\label{app:decomposition}

The decomposition alignment tasks use a deterministic RDKit-based preprocessing step~\citep{rdkit2024} to break each molecule into chemically named building blocks: scaffolds, linkers, functional groups, and substituents.
These structured notes are then fed to a large language model to generate natural-language molecule descriptions.
The supervision for tasks such as \texttt{describe this molecule} is therefore based on a deterministic molecular breakdown and local molecular features, rather than on an unconstrained autoregressive description of the input molecule.
Given a parsed molecule, the pipeline first collects candidate bonds, i.e., bonds that are plausible split points.
These candidates come from five sources: the Murcko scaffold boundary, ring boundaries, the boundary of the union of Ertl functional-group atoms, BRICS retrosynthetic bonds, and boundaries of heteroatom-containing functional-group SMARTS matches.
To avoid cutting through chemically meaningful entities, the pipeline then excludes protected bonds from the cut set: aromatic bonds, bonds inside rings with at most 12 atoms, bonds that would separate exocyclic atoms retained with the ring scaffold, and bonds internal to a functional-group match unless that bond is also a scaffold or ring boundary.
The remaining candidate bonds are cut with \texttt{FragmentOnBonds}; each part keeps its canonical fragment SMILES, original atom indices, and cross-fragment connection atoms.

\begin{figure}[t]
\centering
\includegraphics[width=0.88\textwidth]{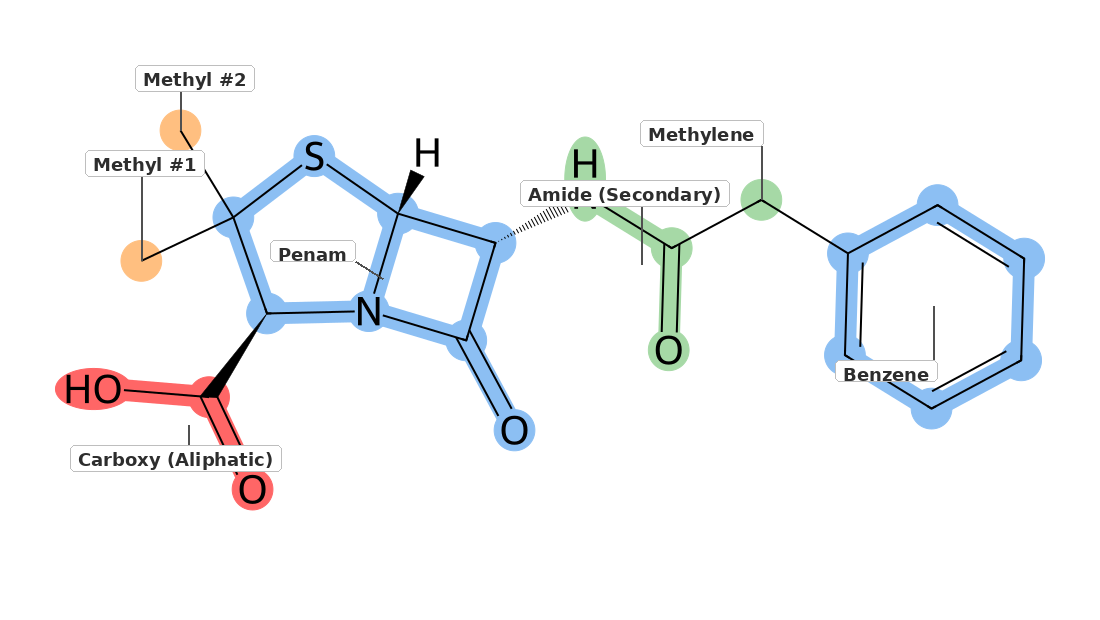}
\caption{Example bond-cutting decomposition for Penicillin G.
Blue highlights ring or ring-system parts, green highlights linker parts, red highlights a functional group, and orange highlights substituents.}
\label{fig:decomposition-penicillin}
\end{figure}

The specification passed to the language model includes a type annotation for every molecular part.
These types make the decomposition readable as chemistry rather than as an unlabelled list of fragments.
To make the cuts sensible, the typing rules use boundary conditions derived from the Murcko scaffold, small rings, macrocycles, Ertl functional groups, and connection count.
Atoms outside the scaffold become functional groups when they are fully covered by an Ertl functional group and otherwise become substituents, or chains when no scaffold exists.
Atoms inside the scaffold become rings, macrocycles, linkers, exocyclic groups, or ring systems depending on whether they are contained in small rings, large rings, or non-ring scaffold paths.
Mixed fragments that cross these boundaries are kept as generic fragments, which exposes failures of the cutting rules instead of silently assigning them to a chemically precise class.

Each typed part is named by a fixed priority cascade: lookup in a local canonical-SMILES structure library, exact match against a local functional-group SMARTS library, single-atom element name with charge label, ring summary by size, aromaticity, heteroatom content, and charge, and finally a carbon-scaffold heuristic such as \texttt{Methyl}, \texttt{Methylene}, or a branched alkyl label.
The structure library is a curated table of common names for frequent ring systems and scaffolds, using many PubChem-derived names but with manual replacements for names that are too verbose or chemically unhelpful.
The SMARTS library is separate from the Ertl functional-group detector: it is a local, ordered set of patterns adapted from RDKit's functional-group hierarchy and extended with custom patterns for decomposition-specific protection and naming.
The formatted notes are ordered by a depth-first traversal that starts from the largest ring-like part and keeps decorations near their parent.
When the molecule contains multiple instances of the same named part, such as two methyl substituents on the same scaffold, the notes distinguish them with \texttt{\#1}, \texttt{\#2}, and so on.
For every part, the notes list the fragment name, type, SMILES, and either ring substituents or non-ring connections; task-specific annotations add only the local feature needed for that alignment task, namely hydrogen-bond acceptors, hydrogen-bond donors, logP contribution, polar surface area, protonation at pH~7.4, stereochemical sites, or Gasteiger partial charges.
Gemini~2.5~Flash receives these structured notes for eight alignment datasets---structure, hydrogen-bond acceptors, hydrogen-bond donors, lipophilicity, polarity, protonation, stereochemistry, and partial charges---and generates the supervised assistant response used in the training mixture.

\subsection{Synthetic CoT Construction}
\label{app:cot-construction}

The downstream panel contains 25 trained TDC binary classification tasks~\citep{huang2021tdc}: AMES, BBB Martins, Bioavailability Ma, HIA Hou, hERG, HIV, ClinTox, skin reaction, M1 agonist, SARS-CoV-2 Touret, four Veith CYP-inhibition tasks, three Carbon-Mangels CYP-substrate tasks, and eight Tox21 assays.
The five held-out endpoints are CYP2D6 Veith, Pgp Broccatelli, PAMPA NCATS, M1 antagonist, and Tox21 PPAR-\(\gamma\).
Each task is trained in direct yes/no and CoT formats.
Training labels are balanced by upsampling the minority class, while validation and test sets keep the TDC scaffold splits unchanged.

CoT targets are synthesized rather than human-written.
The generation prompt contains the canonical SMILES, the same bond-cutting decomposition used in alignment, and the top 20 RDKit descriptors ranked by Gini importance from a 300-tree random forest fit on the task training split.
The ADMET endpoints add the task-specific literature preamble of \autoref{sec:model-architecture}, whose label criterion follows the TDC-linked source paper~\citep{kazius2005toxicophores,pajouhesh2005cns,veber2002bioavailability,veith2009cyp,si2009cyp2c9,carbon2011descriptors,hou2007hia,wu2018moleculenet,broccatelli2011pgp,aronov2005herg}; the Tox21 assays add the assay card instead and require the rationale to follow an adverse-outcome-pathway chain from molecular initiating event to assay outcome.

Both are mined by a Gemini Deep Research agent that surveys the literature for mechanistic priors translating the assay into reasoning over molecular structure, after which Claude Opus 4.6 condenses the report into approximately 600 tokens, to keep training and inference affordable.
The Tox21 assays share one qHTS reporter design, so their reports condense further into a fixed schema of roughly 400 tokens: target family, assay and label definition, biology, ligand chemotypes, and reporter-artifact caveats.
The teacher generates a rationale ending with the task label inside \texttt{<answer>...</answer>} tags: Gemini 3.8 Flash for the ADMET endpoints, Gemini 3.5 Flash for Tox21.
These inputs are used only to construct the supervised CoT targets and are not included in the \bolek{} training prompt.

\begin{figure}[t]
\centering
\includegraphics[width=\textwidth]{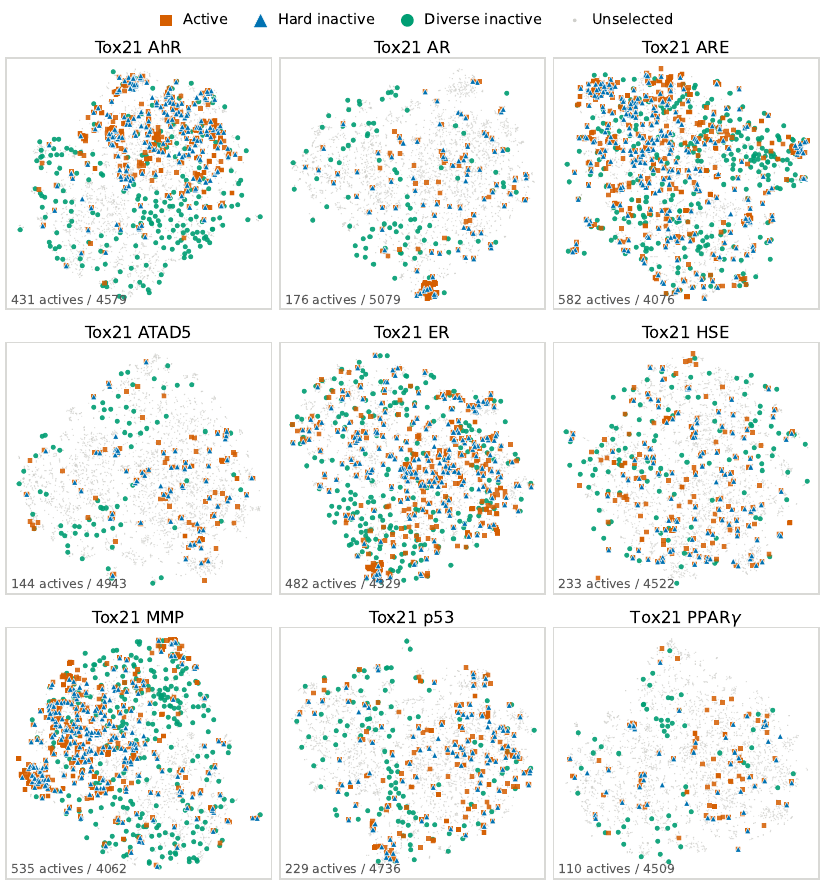}
\caption{Molecules selected for Tox21 CoT generation, in a t-SNE embedding of each assay's training split.
The embedding uses Tanimoto distance between the 2048-bit radius-2 Morgan fingerprints that the selection is computed on.
Hard inactives are the nearest inactive neighbours of an active and sit next to the actives they were matched to.
Diverse inactives are picked by MaxMin from the remaining inactives and also cover regions with few actives.
Grey points are training molecules that receive no CoT.
Tox21 PPAR-\(\gamma\) is held out from training and is shown for completeness.}
\label{fig:tox21-selection}
\end{figure}

Molecule selection differs between the two groups, and is a property of the generation recipe rather than of the endpoint type: the other high-throughput screens in the panel, including HIV, the M1 muscarinic assays, and SARS-CoV-2, follow the ADMET recipe.
For the ADMET endpoints, molecules are deduplicated by InChIKey and keys carrying contradictory labels are dropped, after which the majority class is strided down to at most 40 molecules per minority-class molecule.
Among the endpoints this model trains on, the cap bound only on the M1 agonist screen, whose training split is 0.29\% active; the others were generated in full.
For the Tox21 assays, \texttt{select\_curated} keeps every active plus a matched number of inactives, half of them the Morgan-fingerprint nearest neighbours of an active and half selected by MaxMin for diversity, under a fixed seed.
\autoref{fig:tox21-selection} shows the resulting selection on each assay.

Attempts are generated in one pass rather than conditionally retried, and every attempt whose tagged answer matches the ground-truth label is kept as a separate training row.
The ADMET endpoints use one attempt per molecule, with four further attempts for minority-class molecules on three sparse endpoints.
The Tox21 assays use two label-blind attempts, followed by up to five label-conditioned attempts for molecules that no blind attempt covered, in which the teacher is told the label and asked to derive it.
Those fallback rationales are therefore not label-blind, and their answer match is largely true by construction.
Molecules that yield no matching rationale contribute no CoT training row.

\subsection{Literature-Mined Reasonings}
\label{app:mined-reasoning}

Each TDC task is given an endpoint specification, comprising the permissible answers, the meaning of each label as defined in the source assay publication, a catalogue of mechanisms supporting or opposing each class, and an evidence hierarchy ordered by experimental system, from human measurements down to in silico computations.
For blood--brain-barrier penetration, for instance, the mechanism catalogue contrasts passive transcellular diffusion and LAT1-mediated influx with P-glycoprotein efflux.
An agent based on Gemini 3.8 Flash searches PubMed and produces a schema-validated report in which every hypothesis is a chain of typed steps, for that endpoint the ionisation state, the permeation event, transporter recognition, and the resulting brain partition.
Each step cites its source and carries its own evidence level, and the confidence of a chain is capped by that of its weakest step.
The benchmark label guides the search but is never treated as evidence, and reports contradicting it are retained in a separate subset.
A second model call condenses each report into a four- to eight-sentence rationale tracing the reasoning from chemical structure alone to the report's conclusion.

\subsection{Training Task Examples}
\label{app:training-task-examples}

Tables~\ref{tab:training-examples-alignment}, \ref{tab:training-examples-generation}, \ref{tab:training-examples-decomposition}, and~\ref{tab:training-examples-downstream} show representative supervised examples from the training mixture.
Each row gives the task type, the natural-language prompt, the molecule represented by its SMILES string, and the target assistant answer.
The SMILES column identifies the molecule for readers; in \bolek{} training, the molecule is provided through the \texttt{<molecule>} token and projected molecular vector rather than as prompt text.
The downstream rows use BBB Martins as a representative endpoint, and show the prompt without the assay knowledge of \autoref{sec:model-architecture}, which is injected ahead of the question with 50\% dropout.

\newcommand{\trainingExampleTableSetup}{%
  \tiny
  \setlength{\tabcolsep}{2pt}%
  \renewcommand{\arraystretch}{0.92}%
}

\begin{table}[t]
\centering
\caption{Representative alignment examples for binary, numeric, and list-style tasks.}
\label{tab:training-examples-alignment}
\trainingExampleTableSetup
\begin{tabular}{p{0.16\textwidth}p{0.32\textwidth}p{0.20\textwidth}p{0.24\textwidth}}
\toprule
Task & Question & SMILES & Answer \\
\midrule
FACCS substructure yes/no & In the structure of \texttt{\textless molecule\textgreater}, are there more than three ring bonds? Answer only with yes or no. & \texttt{CC(C)OCCOC(C)C(C)C} & No. \\
MACCS key yes/no & Does N in molecule \texttt{\textless molecule\textgreater} reach O through a chain of three atoms? Answer with yes or no. & \texttt{CC(=O)NC(C)c1ccccn1} & No. \\
RDKit fragment yes/no & Does the structure of \texttt{\textless molecule\textgreater} include a furan? Answer with yes or no. & \texttt{COCC(O)CCNCCCF} & No. \\
RDKit count descriptor & In \texttt{\textless molecule\textgreater}, how many rings are there? Answer with just the number. & \texttt{O=PCCCCCC1CCSS1} & 1. \\
RDKit float descriptor & Tell me the LogP of \texttt{\textless molecule\textgreater}. Answer with just the approximate number. & \texttt{COCCNCC(C)CCCN} & 0.6 \\
Mordred count descriptor & Count the I atoms in \texttt{\textless molecule\textgreater}. Answer with just the number. & \texttt{IC\#CC\#CC\#CC\#CC\#CI} & 2. \\
Mordred float descriptor & For molecule \texttt{\textless molecule\textgreater}, what is the estimated LogS? Answer with just the approximate number. & \texttt{CCOCC(N)c1cscc1C} & -2.4 \\
Mordred ring descriptor & How many heterocyclic rings with twelve members does \texttt{\textless molecule\textgreater} have? Answer with just the number. & \texttt{C1CNCCNCCNCCN1} & 1. \\
SMILES recovery & Provide the SMILES notation of \texttt{\textless molecule\textgreater}. & \texttt{CCCSCC(S)CSCCC} & \texttt{CCCSCC(S)CSCCC} \\
Textbook substructure yes/no & Does molecule \texttt{\textless molecule\textgreater} contain X Fluoride Phenyl? Answer with yes or no. & \texttt{CCC(CCN)Sc1nncs1} & No. \\
Substructure list & List all chemical substructures contained by molecule \texttt{\textless molecule\textgreater}. Do not output anything else except a semicolon-separated list of substructures. & \texttt{CCN(CC(=O)NC)C(C)=O} & \texttt{=O}; Amide \\
\bottomrule
\end{tabular}
\end{table}

\begin{table}[t]
\centering
\caption{Representative free-text generation examples for molecular descriptions and naming tasks.}
\label{tab:training-examples-generation}
\trainingExampleTableSetup
\begin{tabular}{p{0.16\textwidth}p{0.28\textwidth}p{0.25\textwidth}p{0.23\textwidth}}
\toprule
Task & Question & SMILES & Answer \\
\midrule
KnowMol structure description & Provide a clear structural overview of \texttt{\textless molecule\textgreater} and explain how its parts fit together. & \texttt{C1C(C(C(\allowbreak C(C1[NH3+])O)O)O)[NH3+]} & The molecule consists of a cyclohexane ring as the main chain. Attached to the ring are three hydroxyl groups (-OH) at positions 2, 3, and 4. There are two ammonium ions (-NH3+) attached at positions 1 and 5. The hydroxyl groups and ammonium ions are connected directly to the carbon atoms of the cyclohexane ring. The molecule has no side chains, and all functional groups are directly bonded to the ring. The structure includes alkyl groups as part of the cyclohexane ring, hydroxyl groups, and ammonium ions. \\
KnowMol property description & Analyze how the functional groups and overall structure of \texttt{\textless molecule\textgreater} influence its physical and chemical properties, including polarity, acidity or basicity, solubility, reactivity, stereochemistry, and electrophilicity. & \texttt{CCC1=C(N=C(C(=N1)CC)C)C} & The molecule is nonpolar due to its symmetrical structure, which causes the dipole moments to cancel out. It exhibits basicity because of the nitrogen atoms in the pyrazine ring, which can donate electron pairs. The molecule is more soluble in nonpolar solvents due to the presence of nonpolar alkyl groups (ethyl and methyl). Its reactivity is relatively low as it lacks highly reactive functional groups, but the nitrogen atoms can participate in nucleophilic reactions. The molecule does not have chiral centers or double bonds that would lead to stereoisomerism, so it lacks stereochemistry concerns. The electrophilicity is low because the electron-donating alkyl groups reduce the electron deficiency at the nitrogen atoms. \\
ChEBI molecule description & Please describe the molecule \texttt{\textless molecule\textgreater}. & \texttt{C(=O)(N)NS(=O)(=O)NC(=O)N} & The molecule is classified as a sulfuric amide and belongs to the urea family. \\
Molecule naming (docking) & What is the IUPAC name of the compound \texttt{\textless molecule\textgreater}? Write only the name. & \texttt{CCCCCCCCCCC=O} & undecanal \\
Molecule naming (ZINC20) & What systematic name is assigned to the compound \texttt{\textless molecule\textgreater}? Do not output anything else except the name. & \texttt{CCCCCCCCCCC\#CCCO} & tetradec-3-yn-1-ol \\
\bottomrule
\end{tabular}
\end{table}

\begin{table}[t]
\centering
\caption{Representative decomposition examples for local structural and physicochemical descriptions.}
\label{tab:training-examples-decomposition}
\trainingExampleTableSetup
\begin{tabular}{p{0.17\textwidth}p{0.30\textwidth}p{0.22\textwidth}p{0.23\textwidth}}
\toprule
Task & Question & SMILES & Answer \\
\midrule
Decomposition: structure & What are the main structural components of \texttt{\textless molecule\textgreater}? & \texttt{F[C](F)C(F)CC(F)(F)C(F)F} & A five-carbon chain is substituted with seven fluorine atoms. \\
Decomposition: hydrogen-bond acceptors & Where can \texttt{\textless molecule\textgreater} accept hydrogen bonds? & \texttt{CC1CC2CC3(C)C1CCC23} & No hydrogen bond acceptors are present. \\
Decomposition: hydrogen-bond donors & What hydrogen bond donor groups does \texttt{\textless molecule\textgreater} have? & \texttt{COC(C)c1cc(F)ccc1C} & No hydrogen bond donors are present. \\
Decomposition: lipophilicity & Describe the lipophilicity of \texttt{\textless molecule\textgreater}. & \texttt{O=c1ccc2ccnc(=O)c-2co1} & The molecule exhibits a slight hydrophilic character. The two external oxygen atoms present contribute to this property. \\
Decomposition: polarity & Is \texttt{\textless molecule\textgreater} polar or nonpolar? & \texttt{CCC(S)CSC1SCSCS1} & The molecule is nonpolar. \\
Decomposition: protonation & Describe the protonation state of \texttt{\textless molecule\textgreater} at physiological pH. & \texttt{CC1C=NNC1\allowbreak =C1[C+]=CC=[C-]C1} & The molecule is uncharged. \\
Decomposition: stereochemistry & Identify the stereogenic sites in \texttt{\textless molecule\textgreater}. & \texttt{O=S(=O)(CCl)NCc1cc[nH]c1} & There are no stereogenic sites. \\
Decomposition: partial charges & Are there any regions of high or low electron density in \texttt{\textless molecule\textgreater}? & \texttt{CC(C)(C)C=Cc1ccccc1} & Electron density is distributed relatively evenly across this molecule. \\
\bottomrule
\end{tabular}
\end{table}

\begin{table}[t]
\centering
\caption{Representative downstream examples for direct and chain-of-thought binary classification.}
\label{tab:training-examples-downstream}
\trainingExampleTableSetup
\begin{tabular}{p{0.17\textwidth}p{0.31\textwidth}p{0.18\textwidth}p{0.26\textwidth}}
\toprule
Task & Question & SMILES & Answer \\
\midrule
Downstream yes/no classification & Can the molecule \texttt{\textless molecule\textgreater} penetrate the blood-brain barrier? Answer with yes or no. & \texttt{NCCc1cn2ccccc2n1} & No. \\
Downstream chain-of-thought classification & Can the molecule \texttt{\textless molecule\textgreater} penetrate the blood-brain barrier? Start with considering the molecule structure and properties. Place the final answer in \texttt{\textless answer\textgreater...\textless/answer\textgreater} tags, it should be either \texttt{pass} or \texttt{fail}. & \texttt{CNCC(O)\allowbreak c1ccc(O)c(O)c1} & Molecular weight is low (\(\sim\)180), but lipophilicity is low (logP near 0) and polarity is fairly high (TPSA \(\sim\)70). It has many hydrogen-bond donors (4) and multiple acceptors, increasing desolvation cost. The secondary amine is protonated at physiological pH, adding a positive charge that strongly disfavors passive diffusion across the BBB despite the aromatic ring.\texttt{\textless answer\textgreater fail\textless/answer\textgreater} \\
\bottomrule
\end{tabular}
\end{table}

\clearpage
\section{Results}
\label{app:detailed-results}
\subsection{Per-Task Classification Results}
\label{app:per-task}

\autoref{tab:per-task-llm} and \autoref{tab:per-task-supervised} break \autoref{tab:downstream} down into its 30 endpoints.
Every model is scored on the molecules it shares with \bolek{}, so a row compares models on one molecule set; \texttt{papers/bolek/scripts/per\_task\_table.py} regenerates both tables.

\begin{center}
\captionof{table}{Per-task scores for the language models of \autoref{tab:downstream}, scored by ROC AUC on every task. GPT-6-Sol is scored from 5 rollouts per molecule rather than 50, and its yes/no run has no SARS-CoV-2 Touret, which the API refused.}
\label{tab:per-task-llm}
\scriptsize
\resizebox{\textwidth}{!}{%
\begin{tabular}{llcccccccc}
\toprule
Task & Metric & Qwen YN & Qwen CoT & GPT-6-Sol YN & GPT-6-Sol CoT & TxGemma-Chat & TxGemma-Predict & Bolek YN & Bolek CoT \\
\midrule
\multicolumn{10}{l}{\emph{ADMET (17)}} \\
AMES & ROC & 0.583 & 0.700 & 0.721 & 0.790 & 0.680 & 0.776 & 0.793 & 0.807 \\
BBB Martins & ROC & 0.628 & 0.857 & 0.805 & 0.871 & 0.713 & 0.822 & 0.882 & 0.884 \\
Bioavailability Ma & ROC & 0.478 & 0.651 & 0.590 & 0.815 & 0.685 & 0.562 & 0.725 & 0.751 \\
CYP1A2 Veith & ROC & 0.716 & 0.808 & 0.752 & 0.783 & 0.884 & 0.968 & 0.903 & 0.854 \\
CYP2C19 Veith & ROC & 0.614 & 0.637 & 0.719 & 0.760 & 0.829 & 0.963 & 0.851 & 0.784 \\
CYP2C9 Substrate & ROC & 0.591 & 0.550 & 0.639 & 0.705 & 0.632 & 0.600 & 0.604 & 0.630 \\
CYP2C9 Veith & ROC & 0.597 & 0.617 & 0.720 & 0.728 & 0.830 & 0.886 & 0.865 & 0.813 \\
CYP2D6 Substrate & ROC & 0.500 & 0.765 & 0.700 & 0.768 & 0.743 & 0.737 & 0.790 & 0.800 \\
CYP3A4 Substrate & ROC & 0.487 & 0.561 & 0.635 & 0.715 & 0.655 & 0.630 & 0.668 & 0.666 \\
CYP3A4 Veith & ROC & 0.570 & 0.686 & 0.711 & 0.744 & 0.788 & 0.870 & 0.845 & 0.805 \\
hERG & ROC & 0.497 & 0.674 & 0.752 & 0.776 & 0.808 & 0.897 & 0.698 & 0.774 \\
HIA Hou & ROC & 0.663 & 0.891 & 0.930 & 0.983 & 0.906 & 0.892 & 0.962 & 0.984 \\
HIV & ROC & 0.511 & 0.462 & 0.641 & 0.699 & 0.687 & 0.894 & 0.671 & 0.735 \\
ClinTox & ROC & 0.516 & 0.592 & 0.511 & 0.596 & 0.502 & 0.856 & 0.626 & 0.713 \\
M1 agonist & ROC & 0.500 & 0.516 & 0.499 & 0.504 & 0.660 & 0.799 & 0.701 & 0.685 \\
SARS-CoV-2 Touret & ROC & 0.508 & 0.569 & -- & 0.482 & 0.581 & 0.739 & 0.418 & 0.573 \\
Skin reaction & ROC & 0.536 & 0.559 & 0.670 & 0.692 & 0.503 & 0.682 & 0.627 & 0.662 \\
\midrule
\multicolumn{10}{l}{\emph{Tox21 (8)}} \\
Tox21 AhR & ROC & 0.537 & 0.639 & 0.636 & 0.615 & 0.821 & 0.932 & 0.809 & 0.832 \\
Tox21 AR & ROC & 0.495 & 0.487 & 0.507 & 0.536 & 0.721 & 0.832 & 0.762 & 0.702 \\
Tox21 ARE & ROC & 0.481 & 0.548 & 0.565 & 0.615 & 0.795 & 0.903 & 0.758 & 0.747 \\
Tox21 ATAD5 & ROC & 0.502 & 0.507 & 0.582 & 0.587 & 0.695 & 0.932 & 0.704 & 0.763 \\
Tox21 ER & ROC & 0.525 & 0.551 & 0.574 & 0.594 & 0.735 & 0.805 & 0.705 & 0.687 \\
Tox21 HSE & ROC & 0.558 & 0.540 & 0.540 & 0.632 & 0.839 & 0.899 & 0.705 & 0.731 \\
Tox21 p53 & ROC & 0.528 & 0.562 & 0.591 & 0.631 & 0.848 & 0.954 & 0.712 & 0.819 \\
Tox21 MMP & ROC & 0.536 & 0.622 & 0.611 & 0.644 & 0.870 & 0.932 & 0.818 & 0.813 \\
\midrule
\multicolumn{10}{l}{\emph{Held-out (5)}} \\
Pgp Broccatelli & ROC & 0.649 & 0.795 & 0.766 & 0.849 & 0.816 & 0.929 & 0.911 & 0.900 \\
PAMPA NCATS & ROC & 0.661 & 0.648 & 0.726 & 0.673 & 0.583 & 0.689 & 0.708 & 0.726 \\
M1 antagonist & ROC & 0.534 & 0.709 & 0.634 & 0.641 & 0.766 & 0.904 & 0.778 & 0.812 \\
CYP2D6 Veith & ROC & 0.590 & 0.714 & 0.686 & 0.720 & 0.794 & 0.849 & 0.781 & 0.784 \\
Tox21 PPAR-$\gamma$ & ROC & 0.504 & 0.584 & 0.641 & 0.686 & 0.728 & 0.904 & 0.669 & 0.769 \\
\bottomrule
\end{tabular}%
}
\end{center}

\begin{center}
\captionof{table}{Per-task scores for the supervised baselines of \autoref{tab:downstream}, each trained on every endpoint including the five held out from \bolek{}. Metrics follow \autoref{tab:per-task-llm}.}
\label{tab:per-task-supervised}
\scriptsize
\resizebox{\textwidth}{!}{%
\begin{tabular}{llccccc}
\toprule
Task & Metric & Morgan RF & Descriptor RF & Chemprop & Uni-Mol & MolFormer \\
\midrule
\multicolumn{7}{l}{\emph{ADMET (17)}} \\
AMES & ROC & 0.841 & 0.825 & 0.815 & 0.828 & 0.502 \\
BBB Martins & ROC & 0.880 & 0.908 & 0.883 & 0.845 & 0.883 \\
Bioavailability Ma & ROC & 0.699 & 0.712 & 0.572 & 0.685 & 0.664 \\
CYP1A2 Veith & ROC & 0.917 & 0.925 & 0.919 & 0.916 & 0.933 \\
CYP2C19 Veith & ROC & 0.860 & 0.872 & 0.870 & 0.860 & 0.873 \\
CYP2C9 Substrate & ROC & 0.566 & 0.615 & 0.558 & 0.592 & 0.611 \\
CYP2C9 Veith & ROC & 0.863 & 0.870 & 0.868 & 0.863 & 0.860 \\
CYP2D6 Substrate & ROC & 0.795 & 0.807 & 0.825 & 0.724 & 0.763 \\
CYP3A4 Substrate & ROC & 0.617 & 0.666 & 0.603 & 0.656 & 0.591 \\
CYP3A4 Veith & ROC & 0.856 & 0.856 & 0.885 & 0.852 & 0.876 \\
hERG & ROC & 0.815 & 0.829 & 0.737 & 0.762 & 0.813 \\
HIA Hou & ROC & 0.945 & 0.962 & 0.962 & 0.941 & 0.971 \\
HIV & ROC & 0.813 & 0.816 & 0.796 & 0.732 & 0.771 \\
ClinTox & ROC & 0.733 & 0.702 & 0.588 & 0.751 & 0.857 \\
M1 agonist & ROC & 0.645 & 0.679 & 0.676 & 0.579 & 0.709 \\
SARS-CoV-2 Touret & ROC & 0.474 & 0.578 & 0.575 & 0.507 & 0.533 \\
Skin reaction & ROC & 0.648 & 0.652 & 0.591 & 0.595 & 0.604 \\
\midrule
\multicolumn{7}{l}{\emph{Tox21 (8)}} \\
Tox21 AhR & ROC & 0.822 & 0.833 & 0.805 & 0.833 & 0.808 \\
Tox21 AR & ROC & 0.702 & 0.762 & 0.765 & 0.672 & 0.700 \\
Tox21 ARE & ROC & 0.721 & 0.761 & 0.724 & 0.732 & 0.745 \\
Tox21 ATAD5 & ROC & 0.701 & 0.675 & 0.665 & 0.632 & 0.703 \\
Tox21 ER & ROC & 0.713 & 0.786 & 0.704 & 0.717 & 0.723 \\
Tox21 HSE & ROC & 0.651 & 0.784 & 0.688 & 0.714 & 0.587 \\
Tox21 p53 & ROC & 0.776 & 0.806 & 0.789 & 0.718 & 0.668 \\
Tox21 MMP & ROC & 0.800 & 0.879 & 0.833 & 0.837 & 0.843 \\
\midrule
\multicolumn{7}{l}{\emph{Held-out (5)}} \\
Pgp Broccatelli & ROC & 0.920 & 0.901 & 0.909 & 0.871 & 0.913 \\
PAMPA NCATS & ROC & 0.760 & 0.734 & 0.717 & 0.700 & 0.734 \\
M1 antagonist & ROC & 0.824 & 0.831 & 0.828 & 0.742 & 0.828 \\
CYP2D6 Veith & ROC & 0.856 & 0.855 & 0.865 & 0.834 & 0.837 \\
Tox21 PPAR-$\gamma$ & ROC & 0.697 & 0.788 & 0.798 & 0.640 & 0.642 \\
\bottomrule
\end{tabular}%
}
\end{center}

\subsection{Held-Out Regression}
\label{app:heldout-regression}

\autoref{tab:heldout-regression-full} reports the three held-out TDC regression endpoints, which the main text does not discuss.

\begin{center}
\captionof{table}{Held-out TDC regression transfer.
Spearman correlation: higher is better; MAE: lower is better.
For \bolek{}, Lipophilicity and Solubility transfer from descriptor-alignment regression, including LogP- and LogS-like targets, rather than testing fully zero-shot regression.
The \emph{no preamble} column withholds the endpoint's literature preamble at inference.
TxGemma-9B-Chat is included as a specialist reference.
The best score per metric is bolded.}
\label{tab:heldout-regression-full}
\scriptsize
\resizebox{\textwidth}{!}{%
\begin{tabular}{lcccccccc}
\toprule
& \multicolumn{2}{c}{Qwen3-4B-Instruct} & \multicolumn{2}{c}{\bolek{}, no preamble} & \multicolumn{2}{c}{\bolek{}} & \multicolumn{2}{c}{TxGemma-9B-Chat} \\
\cmidrule(lr){2-3}\cmidrule(lr){4-5}\cmidrule(lr){6-7}\cmidrule(lr){8-9}
Task & Spearman & MAE & Spearman & MAE & Spearman & MAE & Spearman & MAE \\
\midrule
Lipophilicity & 0.222 & 2.09 & 0.313 & 1.49 & 0.368 & 1.23 & \textbf{0.418} & \textbf{1.04} \\
PPBR AZ & 0.001 & 15.56 & 0.183 & 25.84 & \textbf{0.340} & 13.73 & -0.001 & \textbf{10.82} \\
Solubility & 0.440 & 2.21 & 0.443 & 2.21 & 0.595 & 1.90 & \textbf{0.768} & \textbf{1.14} \\
\bottomrule
\end{tabular}%
}
\end{center}

\subsection{Rollout-Count Sensitivity}
\label{app:rollout-sensitivity}

This analysis was run early in \bolek{}'s development, to settle how many rollouts an evaluation needs, and is not repeated for the published model.
Its setup and its numbers therefore differ from the rest of the paper: it uses that earlier recipe and its 15 trained endpoints rather than the 30-endpoint panel of \autoref{tab:downstream}.
What it settled is the relative question, how much the rollout count moves a score, and the answer carried forward as the 50-rollout protocol used throughout.

Rollout scoring is a Monte Carlo estimate of the model's positive-answer probability, so a score computed from few generations is noisier than one computed from many.
\autoref{tab:rollout-sensitivity} quantifies that sensitivity by restricting \bolek{} to the first five attempts from the same 50-rollout prediction files.
The mean falls from 0.753 to 0.735, and four of the 15 endpoints score higher at five rollouts than at 50, so the rollout count contributes estimation noise rather than a systematic bias.

\begin{center}
\captionof{table}{Sensitivity of trained-endpoint AUCs to the number of \bolek{} rollouts.
\bolek{} 5-rollout scores use attempts 0--4 from the same prediction files as the 50-rollout column.
Scores keep the TDC task metric, ROC AUC or PR AUC, rather than the ROC AUC used elsewhere in the paper, because this table reuses the earlier runs' prediction files; the best score per row is bolded.}
\label{tab:rollout-sensitivity}
\scriptsize
\resizebox{\textwidth}{!}{%
\begin{tabular}{llcc}
\toprule
Task & Metric & \bolek{} (YN), 50 rollouts & \bolek{} (YN), 5 rollouts \\
\midrule
AMES & ROC AUC & \textbf{0.761} & 0.739 \\
BBB Martins & ROC AUC & \textbf{0.857} & 0.824 \\
Bioavailability Ma & ROC AUC & \textbf{0.774} & 0.712 \\
CYP1A2 Veith & PR AUC & \textbf{0.904} & 0.885 \\
CYP2C19 Veith & ROC AUC & \textbf{0.847} & 0.809 \\
CYP2C9 Substrate & PR AUC & 0.449 & \textbf{0.473} \\
CYP2C9 Veith & PR AUC & 0.716 & \textbf{0.729} \\
CYP2D6 Substrate & PR AUC & 0.607 & \textbf{0.628} \\
CYP2D6 Veith & PR AUC & \textbf{0.612} & 0.584 \\
CYP3A4 Substrate & ROC AUC & \textbf{0.618} & 0.573 \\
CYP3A4 Veith & PR AUC & \textbf{0.792} & 0.775 \\
hERG & ROC AUC & \textbf{0.827} & 0.800 \\
HIA Hou & ROC AUC & 0.934 & \textbf{0.939} \\
HIV & ROC AUC & \textbf{0.679} & 0.665 \\
Pgp Broccatelli & ROC AUC & \textbf{0.923} & 0.894 \\
\midrule
\textbf{Mean} & -- & \textbf{0.753} & 0.735 \\
\bottomrule
\end{tabular}%
}
\end{center}

\subsection{Answer Format per Endpoint}
\label{app:cot-gain}

\begin{figure}[t]
\centering
\includegraphics[width=\linewidth]{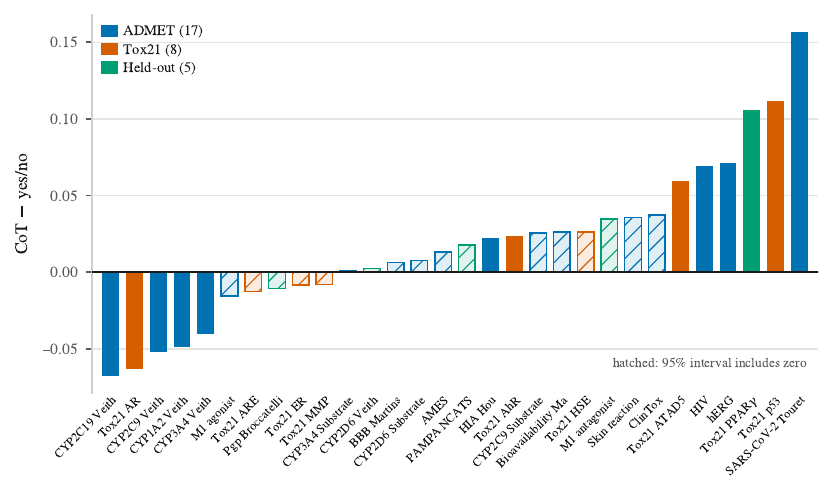}
\caption{Difference between the chain-of-thought and yes/no answer formats on each of the 30 endpoints, ranked.
Each difference carries a 95\% paired bootstrap interval over test molecules.
A hatched bar is one whose interval includes zero, so the difference is not significant; 17 of the 30 are.
Scores are ROC AUC on every task.}
\label{fig:cot-gain}
\end{figure}

\autoref{fig:cot-gain} resolves the task-group means of \autoref{tab:downstream} to the endpoint level, ranked by how much the chain-of-thought format gains over yes/no.

\clearpage
\subsection{TxGemma-9B-Chat Reproduction Comparison}
\label{app:txgemma-comparison}

The TxGemma-9B-Chat scores in the main tables come from our local evaluation of \texttt{google/txgemma-9b-chat}.
For these runs, we use the TDC prompts provided in the model release, extract classification scores from the answer-choice token logits, and parse regression values from the model output.
The TxGemma paper~\citep{wang2025txgemma} reports results from 10-shot prompting, where the prompt includes the most similar examples drawn from training and validation data.
\autoref{tab:txgemma-standard-comparison} therefore compares our direct-question evaluation with the paper's 10-shot setting where a task-level paper value is available, but the two columns should be read as related rather than identical evaluation protocols.
The comparison was run earlier in this work, against the 15 tasks the TxGemma paper reports at task level, so its task set is that of the earlier panel rather than the 30 endpoints of \autoref{tab:downstream}.

\begin{center}
\captionof{table}{TxGemma-9B-Chat direct-question comparison against the TxGemma paper's 10-shot results on the 15 standard downstream tasks.
Metrics follow the TDC task-specific metrics, not the ROC AUC used elsewhere in the paper, since the comparison is against the TxGemma paper's published values.}
\label{tab:txgemma-standard-comparison}
\scriptsize
\begin{tabular}{llccc}
\toprule
Task & Metric & Direct question & 10-shot~\citep{wang2025txgemma} & Difference \\
\midrule
AMES & ROC AUC & 0.680 & 0.723 & $-0.043$ \\
BBB Martins & ROC AUC & 0.717 & 0.832 & $-0.115$ \\
Bioavailability Ma & ROC AUC & 0.685 & 0.666 & $+0.024$ \\
CYP1A2 Veith & PR AUC & 0.886 & 0.862 & $+0.019$ \\
CYP2C19 Veith & ROC AUC & 0.829 & 0.844 & $-0.015$ \\
CYP2C9 Substrate & PR AUC & 0.369 & 0.414 & $-0.045$ \\
CYP2C9 Veith & PR AUC & 0.684 & 0.700 & $-0.016$ \\
CYP2D6 Substrate & PR AUC & 0.560 & 0.653 & $-0.093$ \\
CYP2D6 Veith & PR AUC & 0.461 & 0.517 & $-0.056$ \\
CYP3A4 Substrate & ROC AUC & 0.655 & 0.644 & $+0.011$ \\
CYP3A4 Veith & PR AUC & 0.718 & 0.760 & $-0.042$ \\
hERG & ROC AUC & 0.808 & 0.855 & $-0.047$ \\
HIA Hou & ROC AUC & 0.906 & 0.906 & $0.000$ \\
HIV & ROC AUC & 0.688 & 0.641 & $+0.047$ \\
Pgp Broccatelli & ROC AUC & 0.816 & 0.873 & $-0.057$ \\
\midrule
\textbf{Mean} & -- & 0.698 & 0.726 & $-0.028$ \\
\bottomrule
\end{tabular}
\end{center}

On the standard tasks, our direct-question evaluation is lower on average by 0.028, with four task-level scores higher than the reported 10-shot TxGemma values and one exactly matched score.
The largest gaps are on BBB Martins, CYP2D6 Substrate, CYP2D6 Veith, and Pgp Broccatelli.
The standard-task conclusion is therefore not driven by a single outlier, but exact margins against the paper values should be read with the prompt-format difference in mind.

\subsection{Regression Prediction Scatterplots}
\label{app:regression-scatter}

\begin{figure}[t]
\centering
\includegraphics[width=\textwidth]{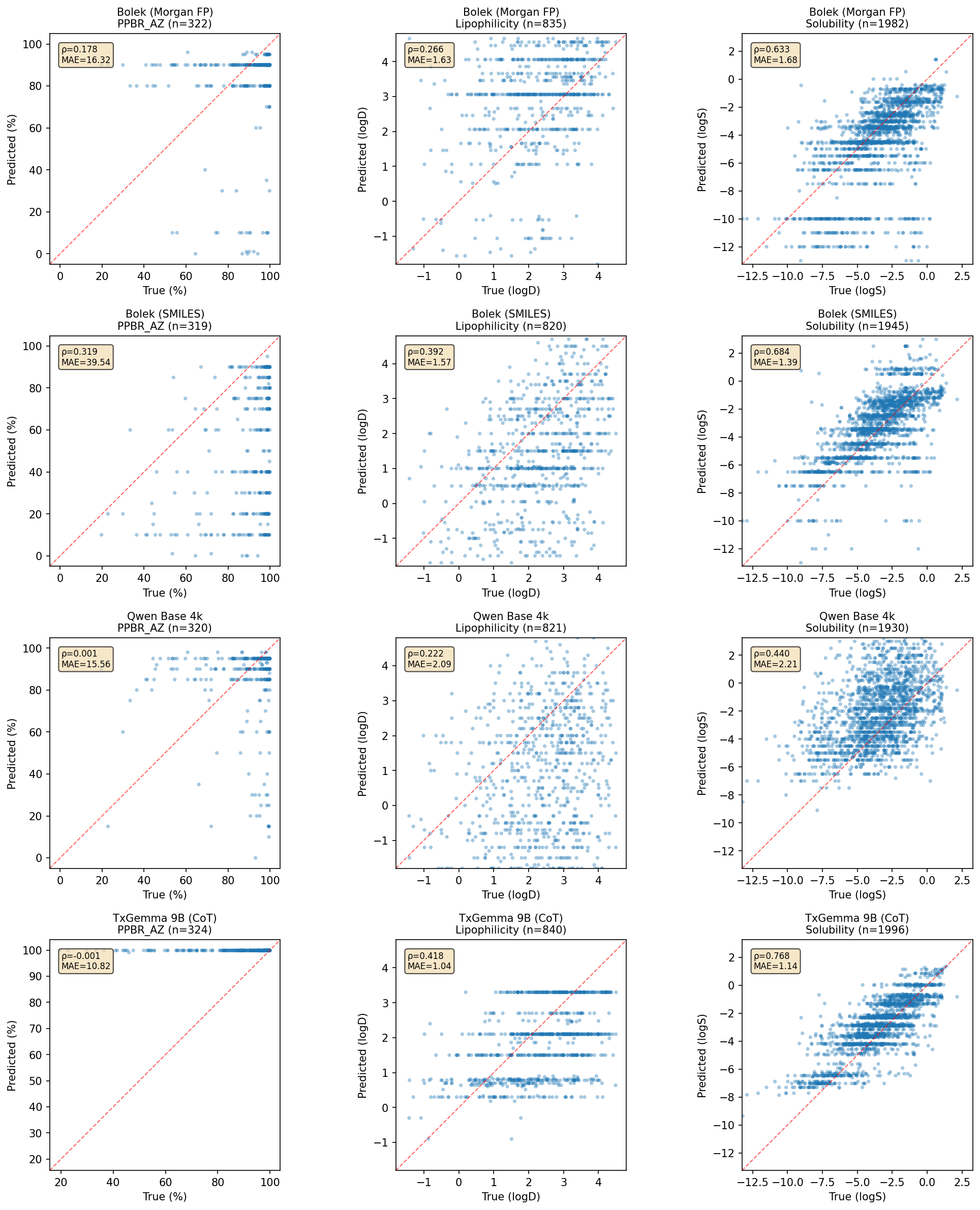}
\caption{Per-molecule held-out regression predictions on PPBR AZ, lipophilicity, and solubility.
Each row is one model, each column is one task, and the dashed line marks perfect calibration.
The inset reports Spearman \(\rho\) and MAE for the corresponding panel.}
\label{fig:regression-scatter-grid}
\end{figure}

\autoref{fig:regression-scatter-grid} shows that the aggregate regression metrics in \autoref{tab:heldout-regression-full} hide qualitatively different prediction behaviors.
On solubility, all models produce a visibly continuous range of values, and TxGemma-9B-Chat is both well ranked and reasonably calibrated relative to the other models.
On PPBR AZ and lipophilicity, however, TxGemma-9B-Chat collapses much of the output space into a small number of repeated values.
For PPBR AZ, it is close to a single-class predictor, with most predictions concentrated near 100\% binding despite broad variation in the ground truth.
For lipophilicity, it behaves more like a few-class classifier than a continuous regressor, assigning predictions to several horizontal bands.

This discretization matters for interpreting the held-out regression table.
TxGemma-9B-Chat has low MAE on PPBR AZ because the endpoint is heavily concentrated near high binding values, but its Spearman correlation is approximately zero and the scatterplot shows little molecule-level ranking signal.
On lipophilicity, its rank correlation remains the best in the table, yet the banded predictions indicate that much of the performance comes from coarse binning rather than calibrated numeric regression.
This pattern is consistent with a task-specific response prior or overfitting to common endpoint ranges, although the scatterplot alone cannot distinguish memorization from a deliberately discretized generation strategy.
\bolek{} and its SMILES variant are less well calibrated in absolute terms, but their predictions remain more continuous on PPBR AZ and solubility, which is useful for rank-ordering molecules under held-out transfer.

\section{Analyses}
\label{app:analyses}
\subsection{Alignment Stage and Supervision}
\label{app:alignment-ablation}

The variants of this block of \autoref{tab:ablations} run the downstream stage of \autoref{tab:training-config} and differ only in its initialization, its trainable parameters, and its data.
Variants without alignment start from Qwen3-4B-Instruct with a newly initialized projector and skip the alignment stage rather than spending its steps on downstream training.
The projector-only variant freezes the language model, embeddings included, and trains 43M of its 4.1B parameters.
Yes/no-only variants drop the chain-of-thought traces and keep the yes/no data at its original relative weights, so they see about 1.5 times as many yes/no examples as \bolek{}.
\autoref{tab:alignment-groups} gives the group-level differences with their intervals, and \autoref{tab:alignment-per-task} breaks them down per task; \texttt{papers/bolek/scripts/alignment\_ablation\_appendix.py} regenerates both.

\begin{table}[t]
\centering
\caption{Differences from \bolek{} in mean ROC AUC for the alignment block of \autoref{tab:ablations}, with 95\% paired bootstrap intervals.
The last column compares the aligned and unaligned yes/no-only variants directly.}
\label{tab:alignment-groups}
\scriptsize
\begin{tabular}{llccccc}
\toprule
& & \multicolumn{4}{c}{Difference from \bolek{}} & Aligned $-$ \\
\cmidrule(lr){3-6}\cmidrule(lr){7-7}
Group & Format & Align., YN & No align. & No align., YN & Projector only & unaligned, YN \\
\midrule
ADMET (17) & CoT & -- & $-0.012$ & -- & $-0.106$ & -- \\
 &  &  & {[}$-0.027$, ${+}0.002${]} &  & {[}$-0.131$, $-0.080${]} &  \\
 & YN & $-0.027$ & $-0.016$ & $-0.060$ & $-0.093$ & ${+}0.033$ \\
 &  & {[}$-0.048$, $-0.007${]} & {[}$-0.036$, ${+}0.003${]} & {[}$-0.083$, $-0.035${]} & {[}$-0.120$, $-0.066${]} & {[}${+}0.013$, ${+}0.052${]} \\
\midrule
Tox21 (8) & CoT & -- & ${+}0.004$ & -- & $-0.094$ & -- \\
 &  &  & {[}$-0.007$, ${+}0.015${]} &  & {[}$-0.112$, $-0.075${]} &  \\
 & YN & $-0.100$ & $-0.018$ & $-0.113$ & $-0.063$ & ${+}0.013$ \\
 &  & {[}$-0.117$, $-0.083${]} & {[}$-0.033$, $-0.001${]} & {[}$-0.130$, $-0.096${]} & {[}$-0.082$, $-0.044${]} & {[}$-0.002$, ${+}0.026${]} \\
\midrule
Held-out (5) & CoT & -- & $-0.011$ & -- & $-0.138$ & -- \\
 &  &  & {[}$-0.027$, ${+}0.006${]} &  & {[}$-0.168$, $-0.107${]} &  \\
 & YN & $-0.130$ & $-0.049$ & $-0.133$ & $-0.131$ & ${+}0.003$ \\
 &  & {[}$-0.156$, $-0.104${]} & {[}$-0.076$, $-0.022${]} & {[}$-0.164$, $-0.102${]} & {[}$-0.164$, $-0.099${]} & {[}$-0.024$, ${+}0.029${]} \\
\bottomrule
\end{tabular}
\end{table}

\begin{table}[t]
\centering
\caption{Per-task difference from \bolek{} in ROC AUC for the alignment block of \autoref{tab:ablations}.
Underlined where the 95\% paired bootstrap interval excludes zero.}
\label{tab:alignment-per-task}
\scriptsize
\begin{tabular}{lcccccc}
\toprule
& \multicolumn{4}{c}{YN} & \multicolumn{2}{c}{CoT} \\
\cmidrule(lr){2-5}\cmidrule(lr){6-7}
Task & Align., YN & No align. & No align., YN & Projector only & No align. & Projector only \\
\midrule
\multicolumn{7}{l}{\emph{ADMET (17)}} \\
AMES & $-0.018$ & \underline{$-0.035$} & \underline{$-0.042$} & \underline{$-0.158$} & \underline{$-0.029$} & \underline{$-0.174$} \\
BBB Martins & \underline{$-0.063$} & $-0.021$ & \underline{$-0.151$} & \underline{$-0.239$} & ${+}0.016$ & \underline{$-0.185$} \\
Bioavailability Ma & $-0.045$ & ${+}0.024$ & \underline{$-0.149$} & $-0.080$ & $-0.018$ & \underline{$-0.140$} \\
CYP1A2 Veith & $-0.007$ & $-0.006$ & \underline{$-0.028$} & \underline{$-0.125$} & \underline{${+}0.014$} & \underline{$-0.065$} \\
CYP2C9 Substrate & $-0.060$ & ${+}0.018$ & ${+}0.037$ & $-0.093$ & $-0.011$ & $-0.090$ \\
CYP2C9 Veith & \underline{$-0.020$} & \underline{$-0.014$} & $-0.002$ & \underline{$-0.098$} & ${+}0.005$ & \underline{$-0.045$} \\
CYP2D6 Substrate & $-0.071$ & $-0.001$ & \underline{$-0.113$} & \underline{$-0.195$} & $-0.004$ & \underline{$-0.203$} \\
CYP2C19 Veith & \underline{${+}0.028$} & $-0.003$ & ${+}0.009$ & \underline{$-0.082$} & ${+}0.008$ & \underline{$-0.026$} \\
CYP3A4 Substrate & ${+}0.034$ & ${+}0.031$ & ${+}0.008$ & $-0.100$ & ${+}0.028$ & $-0.011$ \\
CYP3A4 Veith & ${+}0.011$ & \underline{$-0.013$} & \underline{$-0.015$} & \underline{$-0.067$} & ${+}0.003$ & \underline{$-0.024$} \\
hERG & ${+}0.025$ & ${+}0.069$ & $-0.031$ & $-0.122$ & ${+}0.034$ & \underline{$-0.199$} \\
HIA Hou & \underline{$-0.099$} & $-0.036$ & \underline{$-0.180$} & \underline{$-0.067$} & \underline{$-0.038$} & \underline{$-0.105$} \\
HIV & \underline{$-0.039$} & \underline{${+}0.023$} & \underline{$-0.041$} & \underline{${+}0.047$} & ${+}0.005$ & \underline{$-0.071$} \\
ClinTox & $-0.035$ & $0.000$ & $-0.094$ & $-0.009$ & \underline{$-0.072$} & $-0.047$ \\
M1 agonist & \underline{$-0.144$} & \underline{$-0.146$} & \underline{$-0.106$} & $-0.077$ & $-0.047$ & $-0.096$ \\
SARS-CoV-2 Touret & ${+}0.037$ & $-0.098$ & ${+}0.041$ & ${+}0.069$ & $-0.047$ & $-0.033$ \\
Skin reaction & ${+}0.002$ & $-0.070$ & $-0.163$ & \underline{$-0.183$} & $-0.053$ & \underline{$-0.287$} \\
\midrule
\multicolumn{7}{l}{\emph{Tox21 (8)}} \\
Tox21 AhR & \underline{$-0.115$} & $-0.001$ & \underline{$-0.099$} & \underline{$-0.097$} & $-0.006$ & \underline{$-0.137$} \\
Tox21 AR & \underline{$-0.130$} & \underline{$-0.054$} & \underline{$-0.178$} & \underline{$-0.139$} & ${+}0.002$ & \underline{$-0.120$} \\
Tox21 ARE & \underline{$-0.094$} & $-0.015$ & \underline{$-0.072$} & \underline{$-0.060$} & ${+}0.017$ & \underline{$-0.059$} \\
Tox21 ATAD5 & \underline{$-0.151$} & \underline{$-0.113$} & \underline{$-0.157$} & $-0.021$ & $-0.012$ & \underline{$-0.140$} \\
Tox21 ER & $-0.027$ & ${+}0.029$ & \underline{$-0.057$} & $-0.005$ & ${+}0.022$ & $-0.010$ \\
Tox21 HSE & $-0.065$ & $-0.021$ & \underline{$-0.100$} & $-0.052$ & ${+}0.011$ & \underline{$-0.088$} \\
Tox21 p53 & \underline{$-0.131$} & ${+}0.043$ & \underline{$-0.119$} & $-0.043$ & ${+}0.002$ & \underline{$-0.132$} \\
Tox21 MMP & \underline{$-0.088$} & $-0.009$ & \underline{$-0.124$} & \underline{$-0.088$} & $-0.006$ & \underline{$-0.064$} \\
\midrule
\multicolumn{7}{l}{\emph{Held-out (5)}} \\
CYP2D6 Veith & \underline{$-0.068$} & \underline{$-0.055$} & \underline{$-0.142$} & \underline{$-0.194$} & $-0.001$ & \underline{$-0.148$} \\
Pgp Broccatelli & \underline{$-0.100$} & $-0.022$ & \underline{$-0.116$} & \underline{$-0.151$} & $-0.011$ & \underline{$-0.073$} \\
Tox21 PPAR-$\gamma$ & \underline{$-0.105$} & $-0.033$ & $-0.019$ & ${+}0.026$ & $-0.011$ & \underline{$-0.134$} \\
PAMPA NCATS & \underline{$-0.161$} & ${+}0.008$ & \underline{$-0.114$} & \underline{$-0.119$} & ${+}0.003$ & \underline{$-0.151$} \\
M1 antagonist & \underline{$-0.215$} & \underline{$-0.145$} & \underline{$-0.274$} & \underline{$-0.216$} & $-0.036$ & \underline{$-0.185$} \\
\bottomrule
\end{tabular}
\end{table}

\subsection{Alignment Metrics by Molecular Representation}
\label{app:fp-smiles-alignment}

\autoref{tab:fp-smiles-alignment} compares the four molecular representations of \autoref{tab:ablations} on alignment itself, at the stage-one checkpoint each downstream run continues from.
All four start from Qwen3-4B-Instruct~\citep{yang2025qwen3technicalreport} and follow the same training procedure as each other; only the molecular representation differs.
The fingerprint variant consumes a 2048-bit, radius-2 Morgan fingerprint through the projector, while the SMILES variant consumes the canonical SMILES string in the prompt.

\paragraph{Tasks and metrics.}
Substructure-classification rows (FACCS substructure, MACCS key, RDKit fragment, textbook substructure) ask binary yes/no questions of the form ``does this molecule contain pattern $X$?''~and report accuracy on patterns held out from training (see \emph{Seen vs unseen splits} below).
Substructure list reports the IoU between the semicolon-separated substructure tokens generated by the model and those in the reference list, treating both as multisets.
Molecule naming (docking) and molecule naming (ZINC20) report the IoU between whitespace-separated parts of the predicted IUPAC or systematic name and the reference name.
Descriptor rows report the per-descriptor Pearson correlation $r$ between predicted and ground-truth values, averaged across all descriptors in the family.
The SMILES recovery task is not included: the SMILES variant sees the canonical SMILES directly in the prompt, so emitting it is trivial and the task is dropped from the SMILES training mixture, which makes any FP-versus-SMILES comparison on this row uninformative.

\paragraph{Seen vs unseen splits.}
For each substructure-classification dataset (FACCS, MACCS, RDKit fragment, textbook substructure), we hold out a fixed subset of patterns from training and use them only at evaluation; the complementary patterns are the \emph{seen} subset that appears in training questions.
The unseen rows therefore measure transfer to new patterns rather than memorization of training-time questions, hence we report unseen accuracy.
The remaining alignment tasks (substructure list, molecule naming, descriptor regression) cover the full task space at training time, so they only require a held-out molecule split, taken from the MolPILE validation set.

\begin{table}[t]
\centering
\caption{Alignment metrics for the four molecular representations, at the alignment checkpoint each downstream run continues from.
Substructure-classification rows report accuracy on the unseen pattern split, list and naming rows report token-set IoU, and descriptor rows report Pearson $r$.
The best score per row is bolded.}
\label{tab:fp-smiles-alignment}
\small
\begin{tabular}{lcccc}
\toprule
Metric & Morgan fingerprint & SMILES & Uni-Mol & Molformer-XL \\
\midrule
\multicolumn{5}{l}{\emph{Substructure classification, unseen-pattern accuracy}} \\
FACCS substructure        & 0.890 & \textbf{0.968} & 0.894 & 0.906 \\
MACCS key                 & 0.850 & \textbf{0.900} & 0.826 & 0.871 \\
RDKit fragment            & 0.888 & \textbf{0.939} & 0.844 & 0.876 \\
Textbook substructure     & 0.869 & \textbf{0.941} & 0.829 & 0.872 \\
\emph{Group mean}         & 0.874 & \textbf{0.937} & 0.848 & 0.881 \\
\midrule
\multicolumn{5}{l}{\emph{IoU of substructures / name parts}} \\
Substructure list         & 0.717 & \textbf{0.908} & 0.479 & 0.674 \\
Molecule naming (docking) & 0.362 & \textbf{0.704} & 0.191 & 0.309 \\
Molecule naming (ZINC20)  & 0.184 & \textbf{0.378} & 0.116 & 0.155 \\
\emph{Group mean}         & 0.421 & \textbf{0.663} & 0.262 & 0.379 \\
\midrule
\multicolumn{5}{l}{\emph{Descriptor prediction, Pearson $r$}} \\
RDKit count descriptor    & 0.813 & \textbf{0.960} & 0.759 & 0.871 \\
RDKit float descriptor    & 0.836 & \textbf{0.930} & 0.860 & 0.866 \\
Mordred count descriptor  & 0.724 & \textbf{0.922} & 0.841 & 0.877 \\
Mordred float descriptor  & 0.900 & \textbf{0.979} & 0.922 & 0.907 \\
Mordred ring descriptor   & 0.619 & \textbf{0.834} & 0.600 & 0.757 \\
\emph{Group mean}         & 0.778 & \textbf{0.925} & 0.796 & 0.856 \\
\bottomrule
\end{tabular}
\end{table}

\paragraph{Discussion.}
SMILES leads every alignment row, with the largest gaps on tasks that require reading specific characters, atoms, or counts off the molecule (descriptor regression, substructure list, molecule naming).
This is consistent with SMILES being a more natural representation for an LLM than a 2048-bit hashed fingerprint: the SMILES string exposes atoms, bonds, and ring closures directly to the decoder, while the fingerprint compresses local environments into a fixed-width vector that the projector has to unpack.
Substructure-classification accuracy on unseen patterns also favours SMILES, but by a smaller margin, because the yes/no answer can be derived from coarse fingerprint features.
Among the three vector encoders the ranking depends on the task: Molformer-XL, itself trained on SMILES, is second on substructure classification and on descriptors, while the Morgan fingerprint is second on the IoU tasks and Uni-Mol is last on both.
These gaps do not carry downstream, where the four sit within roughly 0.02 of each other per task group and the fingerprint leads the trained ADMET endpoints (\autoref{tab:ablations}).

\subsection{Backbone Size Ablation}
\label{app:size-ablation}

This ablation was run early in \bolek{}'s development, to settle the choice of backbone, and is not repeated for the published model.
Its setup and its numbers therefore differ from the rest of the paper: the six runs share a joint alignment-plus-downstream mixture trained for 10{,}000 steps rather than the two supervised stages of \autoref{sec:model-architecture}, and they predate both the 25:5 panel and the assay cards.
The columns are comparable with each other but not with \autoref{tab:downstream}, and what they support is the relative question the decision turned on: how performance scales with backbone size, holding the rest of that earlier recipe fixed.

To isolate the effect of backbone size, we re-train \bolek{} on five base Qwen3 checkpoints from 0.6B to 14B parameters, plus the Qwen3-4B-Instruct-2507 backbone used by the main \bolek{} model~\citep{yang2025qwen3technicalreport}.
All six runs share the fingerprint adapter, mixture, optimizer, and schedule, and are evaluated with the same 50-rollout protocol used in the main paper.
The 4B-Instruct column is that earlier recipe's \bolek{} model and serves as the reference point for the rest of this appendix.
A 32B run did not fit on a single 8\,$\times$\,80\,GB H100 node and was not pursued.

Training cost grows roughly linearly with backbone parameters (\autoref{tab:size-cost}).

\begin{table}[h]
\centering
\caption{Approximate H100 training cost per backbone for the size-sweep ablation.}
\label{tab:size-cost}
\small
\begin{tabular}{lccccc}
\toprule
Backbone   & 0.6B & 1.7B & 4B & 8B & 14B \\
\midrule
GPU-hours  & 14   & 18   & 26 & 52 & 102 \\
\bottomrule
\end{tabular}
\end{table}

\textbf{Alignment quality scales monotonically with backbone size.}
\autoref{tab:size-alignment} reports alignment metrics across the sweep, grouped by metric type with a per-group mean.
Substructure-classification accuracy on unseen patterns, IoU of listed substructures and predicted name parts, descriptor Pearson correlation, and SMILES recovery from the fingerprint all rise nearly monotonically from 0.6B to 14B.
The single largest step is between 1.7B and 4B; subsequent doublings still help but at a slower rate.
The 4B-Instruct backbone edges the 4B base backbone on the accuracy, Pearson, and SMILES means but loses to it on the IoU mean, so the instruct boost on alignment is small and not uniform.

\begin{table*}[t]
\centering
\caption{Alignment metrics across backbone sizes.
Substructure-classification rows report unseen-class accuracy, substructure-list and naming rows report IoU between predicted and reference token sets, descriptor rows report Pearson $r$, and SMILES rows report Tanimoto similarity to the canonical SMILES and the fraction of generated SMILES that parse.
The 4B-Instruct column is the backbone used by the main \bolek{} model.
The best score per row is bolded.}
\label{tab:size-alignment}
\scriptsize
\resizebox{\textwidth}{!}{%
\begin{tabular}{lcccccc}
\toprule
Metric & 0.6B & 1.7B & 4B & \shortstack{4B-Instr\\(\bolek{})} & 8B & 14B \\
\midrule
\multicolumn{7}{l}{\emph{Substructure classification, unseen-pattern accuracy}} \\
FACCS substructure        & 0.792 & 0.830 & 0.890 & 0.895 & 0.893 & \textbf{0.908} \\
MACCS key                 & 0.764 & 0.801 & 0.825 & 0.843 & 0.846 & \textbf{0.855} \\
RDKit fragment            & 0.813 & 0.842 & 0.870 & 0.879 & 0.885 & \textbf{0.892} \\
Textbook substructure     & 0.811 & 0.847 & 0.859 & \textbf{0.878} & 0.851 & 0.877 \\
\emph{Group mean}         & 0.792 & 0.824 & 0.855 & 0.868 & 0.863 & \textbf{0.875} \\
\midrule
\multicolumn{7}{l}{\emph{IoU of substructures / name parts}} \\
Substructure list         & 0.638 & 0.690 & 0.726 & 0.709 & \textbf{0.750} & 0.746 \\
Molecule naming (docking) & 0.236 & 0.273 & 0.340 & 0.349 & 0.381 & \textbf{0.401} \\
Molecule naming (ZINC20)  & 0.169 & 0.170 & 0.199 & 0.168 & \textbf{0.201} & 0.191 \\
\emph{Group mean}         & 0.348 & 0.378 & 0.422 & 0.409 & 0.444 & \textbf{0.446} \\
\midrule
\multicolumn{7}{l}{\emph{Descriptor prediction, Pearson $r$}} \\
RDKit count descriptor    & 0.672 & 0.757 & 0.793 & 0.813 & 0.823 & \textbf{0.829} \\
RDKit float descriptor    & 0.645 & 0.713 & 0.841 & 0.846 & 0.861 & \textbf{0.863} \\
Mordred count descriptor  & 0.614 & 0.637 & 0.719 & 0.704 & 0.722 & \textbf{0.745} \\
Mordred float descriptor  & 0.825 & 0.850 & 0.833 & 0.890 & 0.891 & \textbf{0.897} \\
Mordred ring descriptor   & 0.539 & 0.549 & 0.587 & 0.586 & 0.616 & \textbf{0.643} \\
\emph{Group mean}         & 0.659 & 0.701 & 0.755 & 0.768 & 0.783 & \textbf{0.795} \\
\midrule
\multicolumn{7}{l}{\emph{SMILES recovery from fingerprint}} \\
SMILES Tanimoto        & 0.306 & 0.337 & 0.387 & 0.387 & \textbf{0.425} & 0.415 \\
SMILES parsed          & 0.816 & 0.846 & 0.891 & \textbf{0.894} & 0.885 & 0.887 \\
\emph{Group mean}      & 0.561 & 0.592 & 0.639 & 0.641 & \textbf{0.655} & 0.651 \\
\bottomrule
\end{tabular}%
}
\end{table*}

\textbf{Downstream gains plateau from 4B onward.}
\autoref{tab:size-downstream-yn} and \autoref{tab:size-downstream-cot} report ROC or PR AUC across the same six backbones for direct yes/no and chain-of-thought formats.
The yes/no mean improves from 0.708 at 0.6B to 0.749 at 4B, then plateaus: 4B-Instruct reaches 0.753, 8B reaches 0.759, and 14B reaches 0.757.
Chain-of-thought scaling is weaker still, with all backbones from 4B upward within 0.008 of each other on the mean (4B 0.709, 4B-Instruct 0.704, 8B 0.706, 14B 0.712).

A few task-level patterns survive across formats.
HIA Hou is the cleanest positive scaler in yes/no mode, climbing from 0.889 at 0.6B to 0.988 at 14B.
HIV inverse-scales in yes/no mode, dropping from 0.763 at 0.6B to 0.680 at 14B.
The 4B-Instruct backbone wins outright on Bioavailability Ma, CYP2C9 Substrate, and Pgp Broccatelli in yes/no mode and on Bioavailability Ma, BBB Martins, CYP2D6 Substrate, and CYP3A4 Veith in chain-of-thought mode.
However, most of those tasks are small, so the observed wins might be attributed to noisy test sets.

\begin{table*}[t]
\centering
\caption{Downstream yes/no AUC across backbone sizes.
Each entry is estimated from 50 rollouts per molecule.
Scores keep the TDC task metric, ROC AUC or PR AUC, rather than the ROC AUC used elsewhere in the paper, because this sweep predates it.
The best score per row is bolded.}
\label{tab:size-downstream-yn}
\scriptsize
\resizebox{\textwidth}{!}{%
\begin{tabular}{llcccccc}
\toprule
Task & Metric & 0.6B & 1.7B & 4B & \shortstack{4B-Instr\\(\bolek{})} & 8B & 14B \\
\midrule
AMES               & ROC AUC & 0.640 & 0.712 & 0.740 & 0.761 & \textbf{0.786} & \textbf{0.786} \\
BBB Martins        & ROC AUC & 0.785 & 0.802 & 0.841 & 0.857 & \textbf{0.869} & 0.864 \\
Bioavailability Ma & ROC AUC & 0.689 & 0.658 & 0.737 & \textbf{0.774} & 0.731 & 0.679 \\
CYP1A2 Veith       & PR AUC  & 0.860 & 0.891 & 0.904 & 0.904 & 0.907 & \textbf{0.909} \\
CYP2C19 Veith      & ROC AUC & 0.824 & 0.847 & 0.860 & 0.847 & 0.867 & \textbf{0.872} \\
CYP2C9 Substrate   & PR AUC  & 0.330 & 0.347 & 0.427 & \textbf{0.449} & 0.374 & 0.402 \\
CYP2C9 Veith       & PR AUC  & 0.674 & 0.715 & 0.745 & 0.716 & 0.748 & \textbf{0.763} \\
CYP2D6 Substrate   & PR AUC  & 0.599 & 0.594 & \textbf{0.678} & 0.607 & 0.605 & 0.621 \\
CYP2D6 Veith       & PR AUC  & 0.563 & 0.591 & 0.593 & 0.612 & \textbf{0.632} & 0.615 \\
CYP3A4 Substrate   & ROC AUC & 0.620 & \textbf{0.676} & 0.573 & 0.618 & 0.675 & 0.627 \\
CYP3A4 Veith       & PR AUC  & 0.749 & 0.778 & 0.815 & 0.792 & 0.819 & \textbf{0.827} \\
hERG               & ROC AUC & 0.804 & 0.780 & 0.780 & 0.827 & \textbf{0.844} & 0.802 \\
HIA Hou            & ROC AUC & 0.889 & 0.871 & 0.928 & 0.934 & 0.955 & \textbf{0.988} \\
HIV                & ROC AUC & \textbf{0.763} & 0.704 & 0.705 & 0.679 & 0.664 & 0.680 \\
Pgp Broccatelli    & ROC AUC & 0.836 & 0.906 & 0.907 & \textbf{0.923} & 0.911 & 0.914 \\
\midrule
\textbf{Mean}      & --      & 0.708 & 0.725 & 0.749 & 0.753 & \textbf{0.759} & 0.757 \\
\bottomrule
\end{tabular}%
}
\end{table*}

\begin{table*}[t]
\centering
\caption{Downstream chain-of-thought AUC across backbone sizes.
Conventions match \autoref{tab:size-downstream-yn}.}
\label{tab:size-downstream-cot}
\scriptsize
\resizebox{\textwidth}{!}{%
\begin{tabular}{llcccccc}
\toprule
Task & Metric & 0.6B & 1.7B & 4B & \shortstack{4B-Instr\\(\bolek{})} & 8B & 14B \\
\midrule
AMES               & ROC AUC & 0.652 & 0.667 & 0.714 & 0.729 & 0.726 & \textbf{0.741} \\
BBB Martins        & ROC AUC & 0.784 & 0.809 & 0.825 & \textbf{0.852} & 0.808 & 0.819 \\
Bioavailability Ma & ROC AUC & 0.728 & 0.692 & 0.687 & \textbf{0.730} & 0.729 & 0.668 \\
CYP1A2 Veith       & PR AUC  & 0.859 & 0.868 & 0.874 & \textbf{0.876} & \textbf{0.876} & 0.871 \\
CYP2C19 Veith      & ROC AUC & \textbf{0.774} & 0.755 & 0.728 & 0.713 & 0.717 & 0.731 \\
CYP2C9 Substrate   & PR AUC  & 0.387 & 0.364 & \textbf{0.489} & 0.416 & 0.394 & 0.434 \\
CYP2C9 Veith       & PR AUC  & 0.614 & \textbf{0.616} & 0.607 & 0.579 & 0.580 & 0.601 \\
CYP2D6 Substrate   & PR AUC  & 0.610 & 0.618 & 0.651 & \textbf{0.668} & 0.663 & 0.649 \\
CYP2D6 Veith       & PR AUC  & 0.477 & 0.503 & 0.555 & 0.551 & 0.556 & \textbf{0.560} \\
CYP3A4 Substrate   & ROC AUC & 0.599 & 0.606 & 0.626 & 0.623 & 0.618 & \textbf{0.665} \\
CYP3A4 Veith       & PR AUC  & 0.722 & 0.718 & 0.728 & \textbf{0.737} & 0.724 & 0.733 \\
hERG               & ROC AUC & 0.808 & 0.808 & 0.827 & 0.779 & \textbf{0.837} & \textbf{0.837} \\
HIA Hou            & ROC AUC & 0.888 & 0.860 & 0.888 & 0.869 & \textbf{0.929} & 0.928 \\
HIV                & ROC AUC & 0.533 & \textbf{0.535} & 0.515 & 0.509 & 0.531 & 0.528 \\
Pgp Broccatelli    & ROC AUC & 0.889 & 0.896 & 0.919 & \textbf{0.923} & 0.896 & 0.916 \\
\midrule
\textbf{Mean}      & --      & 0.688 & 0.688 & 0.709 & 0.704 & 0.706 & \textbf{0.712} \\
\bottomrule
\end{tabular}%
}
\end{table*}

The sweep supports three observations:
\begin{itemize}
  \item \textbf{Alignment quality scales monotonically with backbone size}, so the largest backbone we tested is the strongest on alignment.
  \item \textbf{Downstream gains plateau from 4B onward}: doubling parameters from 4B to 8B improves the yes/no mean by 1.0\,pp (0.749 to 0.759), and a further doubling to 14B does not improve it (0.757); chain-of-thought scaling is weaker still, with 14B only 0.3\,pp above 4B on the mean.
  \item \textbf{Training cost grows roughly linearly with backbone size}, so a 14B run uses about four times the H100-hours of a 4B run for an essentially flat downstream return.
\end{itemize}
The 4B-Instruct checkpoint adds a small additional boost over the 4B base on top of these trends, plausibly because the joint alignment and downstream mixture already includes a large body of chemical instruction-tuning supervision and partly overrides whatever instruction priors Qwen3-4B-Instruct contributes. \textbf{Together, these observations motivate Qwen3-4B-Instruct as the main backbone for \bolek{}.}

\subsection{Counterfactual Rationale Test}
\label{app:counterfactual}

A chain-of-thought answer is worth auditing only if the rationale actually guides it.
\bolek{} could instead answer from the molecular token and treat the rationale as a distraction written alongside a decision it has already made.
Correcting such a rationale would then leave the prediction unchanged.
We test which is the case by replacing the model's rationale with one written for a different molecule.

For each test molecule we keep the molecular token and the question fixed.
We then replace the model's own rationale with a \emph{donor rationale} that \bolek{} generated for another test molecule of the same endpoint.
Donors come only from rollouts that reached the correct label for their own molecule, so each donor argues a known label.
We remove the donor's final verdict sentence and discard any donor that still names a label.
The donor argues either the target's true label or the opposite one.
Each molecule receives five donors per condition.
We score both answer labels from the model's log-probabilities after the donor rationale, average the positive-label probability over the five donors, and predict positive at 0.5.
The analysis covers the full test splits of all 30 endpoints in \autoref{tab:downstream} under the default protocol, so both the scored prompt and the donor rollouts carry the literature preamble or assay card.

\begin{table}[t]
\centering
\caption{Counterfactual rationale test on \bolek{}, macro-averaged over the endpoints of each task group.
The model's rationale is replaced by a verdict-stripped rationale that \bolek{} wrote for another molecule of the same endpoint, arguing either the target's true label or the opposite one.
Accuracy uses the mean positive-label probability over five donors per molecule.
Donor consistency is the share of single donors that the model agrees with.}
\label{tab:counterfactual}
\footnotesize
\begin{tabular}{lcccc}
\toprule
& \multicolumn{2}{c}{Accuracy} & \multicolumn{2}{c}{Donor consistency} \\
\cmidrule(lr){2-3}\cmidrule(lr){4-5}
Task group & True label & Opposite & True label & Opposite \\
\midrule
ADMET (17) & 99.8\% & 2.1\% & 97.4\% & 94.0\% \\
Tox21 (8) & 99.6\% & 1.8\% & 98.4\% & 93.0\% \\
Held-out (5) & 100.0\% & 0.6\% & 98.5\% & 94.4\% \\
All (30) & 99.8\% & 1.8\% & 97.9\% & 93.8\% \\
\bottomrule
\end{tabular}
\end{table}

The answer follows the donor rationale almost without exception (\autoref{tab:counterfactual}).
Accuracy is 99.8\% when the donor argues the true label and 1.8\% when it argues the opposite one, and ROC AUC is 1.000 and 0.000 in every task group.
With a true-label donor no endpoint falls below 97.2\%.
With an opposite donor only M1 agonist rises above 12.6\%, to 30.7\%, because the model follows rationales arguing agonism for only about 60\% of molecules.
The molecular token and the question are unchanged and the verdict is removed, so the rationale, rather than the molecule alone, decides the answer.

\section{Licenses and Terms of Use for External Assets}
\label{app:licenses}
\begin{center}
\small
\setlength{\tabcolsep}{3pt}
\begin{tabular}{p{3.2cm}p{1.6cm}p{3.5cm}p{4cm}}
\toprule
Asset & Type & License / Terms & Use in this work \\
\midrule
Qwen3-4B-Instruct & Model & Apache 2.0 & SFT base model \\
TxGemma-9B Chat and Predict & Model & Health AI Dev.\ Foundations ToU (Gemma 2 lineage) & Frozen eval baselines \\
GPT-6-Sol & API & OpenAI Terms of Use & Frozen eval baseline \\
Uni-Mol & Model & MIT & Representation ablation, eval baseline \\
Molformer-XL & Model & Apache 2.0 & Representation ablation, eval baseline \\
Chemprop & Code & MIT & Eval baseline \\
Gemini 2.5 Flash & API & Google Generative AI ToS & Per-part decomposition supervision \\
Gemini 3.5 / 3.8 Flash & API & Google Generative AI ToS & Synthetic CoT generation \\
Gemini 3.8 Flash & API & Google Generative AI ToS & Literature mining and rationale distillation \\
Claude Opus 4.6 & API & Anthropic Usage Policies & Literature preamble generation \\
TDC (codebase) & Code & MIT & Task definitions, scaffold splits \\
TDC (datasets) & Data & Per-dataset (see TDC repo) & Train/eval molecules and labels \\
MoleculeNet & Data & MIT via DeepChem & Lipophilicity regression \\
AqSolDB & Data & CC0 & Solubility regression \\
ZINC20 & Data & Free to use; terms unspecified & Alignment molecules \\
PubChem (names) & Data & U.S.\ public domain & Name-prediction supervision \\
NCI AIDS Antiviral & Data & U.S.\ public domain & HIV classification \\
MolPILE & Data & MIT & Alignment molecules \\
KnowMol & Data & MIT & Multi-level annotations \\
ChEBI-20-MM & Data & MIT & Molecule descriptions \\
SureChEMBL alerts & Data & CC BY 4.0 & Substructure vocabulary \\
SMARTS-RX & Data & CC BY 4.0 & Substructure vocabulary \\
RCSB PDB~\citep{Berman2000} & Data & CC0 1.0 Public Domain Dedication & Name-prediction supervision \\
RDKit & Code & BSD 3-Clause & Descriptors, fingerprints \\
Mordred & Code & BSD 3-Clause & Descriptors \\
\bottomrule
\end{tabular}
\captionof{table}{External assets used in this work, with licenses and terms of use.}
\label{tab:external-asset-licenses}
\end{center}

\end{document}